\documentclass[lettersize,journal]{IEEEtran}
\usepackage{cite}
\usepackage{amsmath,amssymb,amsfonts}
\usepackage{bbm}
\usepackage{graphicx}
\graphicspath{{./figures/}}
\DeclareGraphicsExtensions{.pdf,.jpeg,.png}
\usepackage{textcomp}
\usepackage{xcolor}
\usepackage{units}
\usepackage{stfloats}
\usepackage{subfigure}
\usepackage{multirow}
\usepackage{booktabs}
\usepackage[inline]{enumitem}
\usepackage[hyphens]{url}
\usepackage[bookmarks=true]{hyperref}
\usepackage{array}
\usepackage{tablefootnote}

\usepackage[linesnumbered,ruled,vlined]{algorithm2e}
\SetKwInput{KwInput}{Input}                
\SetKwInput{KwOutput}{Output}              
\DeclareMathOperator{\atan2}{atan2}

\newcommand{\ie}{\textit{i.e.,}\ }
\newcommand{\eg}{\textit{e.g.,}\ } 

\begin{document}
\title{DRL-VO: Learning to Navigate Through Crowded Dynamic Scenes Using Velocity Obstacles}

\author{Zhanteng Xie,~\IEEEmembership{Student Member,~IEEE}, and Philip Dames,~\IEEEmembership{Member,~IEEE}
\thanks{*This work was funded by the Amazon Research Awards Program, NSF grant IIS-1830419, and Temple University.}
\thanks{Zhanteng Xie and Philip Dames are with the Department of Mechanical Engineering,
        Temple University, Philadelphia, PA, USA
        {\tt\small \{zhanteng.xie, pdames\}@temple.edu}}%
}

\maketitle

\begin{abstract}
This paper proposes a novel learning-based control policy with strong generalizability to new environments that enables a mobile robot to navigate autonomously through spaces filled with both static obstacles and dense crowds of pedestrians. The policy uses a unique combination of input data to generate the desired steering angle and forward velocity: a short history of lidar data, kinematic data about nearby pedestrians, and a sub-goal point. The policy is trained in a reinforcement learning setting using a reward function that contains a novel term based on velocity obstacles to guide the robot to actively avoid pedestrians and move towards the goal. Through a series of 3D simulated experiments with up to 55 pedestrians, this control policy is able to achieve a better balance between collision avoidance and speed (\ie higher success rate and faster average speed) than state-of-the-art model-based and learning-based policies, and it also generalizes better to different crowd sizes and unseen environments. An extensive series of hardware experiments demonstrate the ability of this policy to directly work in different real-world environments with different crowd sizes with zero retraining. Furthermore, a series of simulated and hardware experiments show that the control policy also works in highly constrained static environments on a different robot platform without any additional training. Lastly, several important lessons that can be applied to other robot learning systems are summarized. Multimedia demonstrations are available at \url{https://www.youtube.com/watch?v=KneELRT8GzU&list=PLouWbAcP4zIvPgaARrV223lf2eiSR-eSS}.
\end{abstract}


\section{Introduction}
\label{sec:introduction}
\begin{figure}[t]
    \centering
    \subfigure[Simulated indoor environment]
    {
        \centering
        \includegraphics[width=0.77\linewidth]{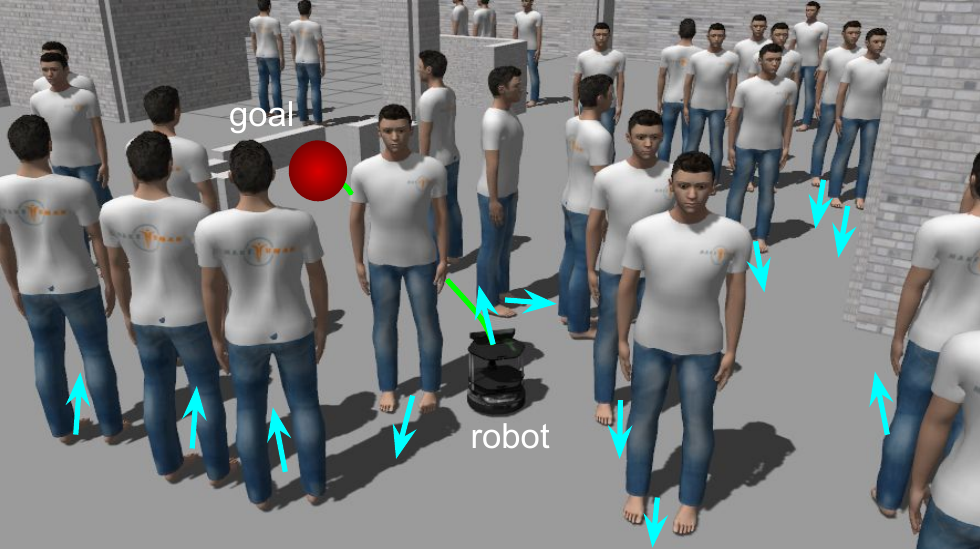}
        \label{fig:gazebo_navigation}
    } 
    \vspace{0.01cm}
    \subfigure[Real-world outdoor environment]
    {
        \centering
        \includegraphics[width=0.77\linewidth]{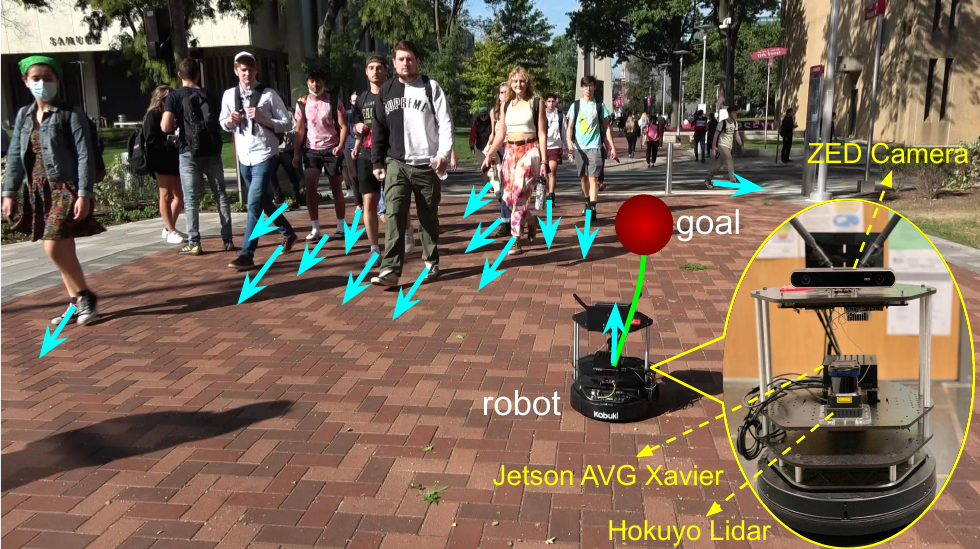}
        \label{fig:real_world_navigation}
    }%
     \caption{
     Our robot navigates through crowds in a simulated indoor environment and a real-world outdoor environment. 
     The robot is following a nominal path (green line) to the goal (red disk) while avoiding pedestrians (blue arrows show current velocity).
    }
    \label{fig:navigation_examples}
\end{figure}

\IEEEPARstart{O}{ne} common application of autonomous mobile robots is replacing manual labor to provide last-mile delivery services.
For example, delivering sterile supplies and injection medicines to patients in hospitals, delivering materials to various packaging workstations in warehouses, and delivering delicious food or groceries to customers in restaurants and grocery stores \cite{Keimyung2021:online,  OneBig2012:online, DemandFo59:online}. 
All these tasks have time limits and require mobile robots to navigate autonomously and quickly to destinations through a partially known space filled with moving people and other static obstacles, as shown in Fig.~\ref{fig:navigation_examples}. 
The main challenges faced by these mobile robots are perceiving complex environments, especially unknown and dynamic pedestrians; extracting useful information; and generating a policy that yields autonomous navigation.  


In this paper, we use a novel deep reinforcement learning-based (DRL) approach due to the complexity of the environments we want the robot to operate in, with 10's of pedestrians moving in dense crowds, which makes it difficult to design clear rules within a traditional model-based framework or to collect sufficiently representative training data to apply supervised learning techniques.
While the use of DRL to generate control policies is becoming commonplace, this paper presents eight primary contributions:
\begin{enumerate*}[label=\arabic*)]
    \item We create a novel combination of data representations for the input to our neural network-based control policy, using a short history of pooled lidar data, the current kinematics of nearby pedestrians, and a sub-goal point. 
    We design these representations to be robust to localization errors by putting all data into the robot's local coordinate frame and to improve generalizability to new environments by handcrafting higher-level representations from the raw sensor data.
    
    \item We apply the theory of velocity obstacles to create a new reward function for the DRL framework that encourages the robot to proactively avoid collisions.
    
    \item We validate the navigation performance in multiple simulated 3D environments and demonstrate that the proposed control policy achieves a better trade-off between collision avoidance and speed, and generalizes better to different crowd sizes and new unseen environments than state-of-art approaches, including a model-based controller \cite{fox1997dynamic}, a supervised learning-based approach \cite{xie2021towards}, and two DRL-based approaches \cite{guldenring2020learning}.
    
    \item We demonstrate that our policy, trained only within a single simulated environment with a single crowd size, can be directly used on a real-world robot in unseen environments, including both indoor and outdoor locations on Temple's campus, and with different crowd densities without any retraining.
    
    \item We demonstrate that our proposed control policy works effectively in highly constrained static environments and on different robot platforms without any retraining. To do this, we competed in the Benchmark Autonomous Robot Navigation (BARN) Challenge at the 2022 IEEE International Conference on Robotics and Automation (ICRA), where we placed 1st in the simulation competition and 3rd in the hardware competition \cite{xiao2022autonomous}. 
    
    \item We make our code and our 3D human-robot interaction simulator available open source at: \url{https://github.com/TempleRAIL/drl_vo_nav}.
    
    \item We conduct a detailed literature review for the most closely related works ranging from model-based to learning-based approaches. 
    
    \item We summarize several important design principles from our work that can be applied to other robot learning systems.
\end{enumerate*}

\section{Related Works}
\label{sec:related_works}
In this section, we provide a detailed description of prior work on robot navigation problems, which we divide into two categories: model-based and learning-based. 
Learning-based approaches can be further divided into supervised learning-based and reinforcement learning-based.

\subsection{Model-based Approaches}
\label{subsec:model_based_approahes}
A typical model-based approach is the ROS \cite{quigley2009ros} navigation stack, which uses costmaps to store obstacle information and uses the dynamic window approach (DWA) planner \cite{fox1997dynamic} to do local planning.
Based on the ROS navigation stack, \cite{singamaneni2021human} recently add human safety and visibility costmaps into both global and local planners to handle human-aware navigation problems.
Another class of model-based approaches are velocity obstacle-based algorithms \cite{fiorini1998motion, wilkie2009generalized, van2011reciprocal}, which map the dynamic environment into the robot velocity space and generate safe control velocities from these velocity constraints. 
Similarly, Arul et al.~\cite{arul2021v} combine reciprocal velocity obstacles with buffered Voronoi cells to solve multi-agent navigation problems.
A third class of model-based approaches uses model predictive control \cite{brito2019model, shen2020collision}, which can integrate collision avoidance and obstacle dynamics into robot constraints. 
Other model-based approaches attempt to directly model how humans and robots interact, using social forces \cite{helbing1995social, ferrer2013robot}, Gaussian mixture models (GMM) \cite{sebastian2017socially}, or a combination of potential functions and limit cycles \cite{boldrer2020socially}.
All those model-based methods require knowledge of pedestrian kinematics, utilize multi-stage procedures to process sensor data, and often require practitioners to carefully hand-tune model parameters, making them difficult to implement and generalize to new scenarios.

\subsection{Supervised Learning-based Approaches}
\label{subsec:supervised_learning_based_approahes}
With the success of deep learning in the computer vision area, many researchers started to apply learning-based approaches to other problems, including robot navigation.
Unlike the traditional model-based approaches with multi-stage procedures, several recent works use deep neural networks to learn end-to-end control policies that generate steering commands directly from raw sensor data.
Pfeiffer et al.~\cite{pfeiffer2017perception} create a data-driven end-to-end motion planner that uses a convolutional neural network (CNN) to generate a linear and angular velocity from raw lidar data. 
Tai et al.~\cite{tai2016deep} also use a CNN to select from one of five discrete robot control commands using raw depth camera data as input.
Loquercio et al.~\cite{loquercio2018dronet} use a similar approach with raw camera image data to train a control policy that enables a drone to drive safely in the streets. 
Similarly, Kahn et al.~\cite{kahn2021badgr} also use the raw camera image data to train an end-to-end learning-based mobile robot navigation system that can navigate in real-world environments with geometrically distracting obstacles.
However, most of these end-to-end approaches only focus on static environments.

For dynamic environments, Pokle et al.~\cite{pokle2019deep} use multiple CNN networks to learn multimodal high-level features from raw sensor data. A CNN-based local planner fuses these features to generate velocity commands.
However, their learned feature-based policy was only tested in a relatively open simulated environment with no more than 3 moving pedestrians.
For highly crowded dynamic environments with up to 50 moving pedestrians, our previous work \cite{xie2021towards} proposes a CNN-based policy that combined a short history of lidar data with kinematic data about nearby pedestrians.\footnote{We utilize a very similar network architecture in this current paper.}
However, as we will show, the policy trained in a supervised setting is not as successful as the approach proposed in this work, particularly when applied to new environments or crowd sizes.

\subsection{Reinforcement Learning-based Approaches}
\label{subsec:reinforcement_learning_based_approahes}
Compared with supervised learning-based methods, DRL-based approaches are more commonly used for navigation in dynamic environments.
Using the proximal policy optimization (PPO) algorithm \cite{schulman2017proximal}, Long et al.~\cite{long2018towards,fan2020distributed} first propose a fully decentralized multi-robot collision avoidance framework using 2D raw lidar range data. 
Following this line of thought, Guldenring et al.~\cite{guldenring2020learning} investigate the different state representations of 2D lidar data, and use the PPO training algorithm to train a human-aware navigation policy. 
Arpino et al.~\cite{perez2020robot} present a multi-layout training regime that uses raw lidar data as its input to train an indoor navigation policy. 
Huang et al.~\cite{huang2021towards} construct a multi-modal late fusion network to fuse raw lidar data and raw camera image data.
Although these sensor-level approaches can quickly and easily generate control policies, they cannot distinguish between static or dynamic obstacles, nor can they utilize high-level information such as the positions and velocities of obstacles.

To address the limitations of using only raw sensor data, many other works leverage information about the pedestrian motion to enable safe navigation in dynamic environments.
Everett et al.~\cite{chen2017socially,everett2018motion,everett2021collision} directly feed the relative position and velocity of pedestrians into the neural network, and generate socially aware collision avoidance policies for pedestrian-rich environments.  
Chen et al.~\cite{chen2019crowd} focus on jointly modeling human-robot and human-human interactions by a self-attention mechanism. 
Immediately afterwards, they also propose a relational graph learning network to infer robot-human interactions and predict their trajectories \cite{chen2020relational}.   
Similarly, another work~\cite{chen2020robot} encodes the crowd state with a graph convolutional network (GCN) and predicts human attention. 
Recently, Liu et al.~\cite{liu2020decentralized} use a decentralized structural-recurrent neural network (RNN) to model the interaction between the robot and pedestrians. 
One major drawback of these agent-level approaches is that they are only suitable for solving crowded navigation problems in open spaces and ignore static obstacles and geometric constraints in the layout.  
To solve this issue, Liu et al.~\cite{liu2020robot} expand the work of \cite{chen2019crowd} by adding additional static obstacle maps, but they need two parallel network models to handle the presence/absence of pedestrians. 
Sathyamoorthy et al.~\cite{sathyamoorthy2020densecavoid} expand the PPO training policy  of \cite{long2018towards} by using a late fusion network to combine raw lidar data with pedestrian trajectory predictions.
Dugas et al.~\cite{dugas2020navrep} use unsupervised learning architectures to predict and reconstruct future lidar latent states.
However, most of the above-described works only focus on exploiting the sensor perception information and its data representations, but ignore the reward function design. 

Towards this, Xu et al.~\cite{xu2021human} propose a knowledge distillation reward term to encourage the robot to learn expert behaviors.  
Patel et al.~\cite{pateldwa} propose a square warning region-based reward function to encourage the robot to navigate in the direction opposite to the heading direction of obstacles. 
However, the effect of knowledge distillation reward largely depends on the quality of the expert dataset, and the reward function based on the square warning region is too conservative, resulting in slow actions.

\subsection{Comparative Analysis}
\label{subsec:summary_analysis}
There is an abundance of studies that focus on robot navigation problems, with solutions ranging from traditional model-based approaches to supervised learning-based approaches to reinforcement learning-based approaches.
Typical model-based approaches compute efficient paths and the parameters are easily interpretable, but require manually adjusting model parameters for different scenarios \cite{fox1997dynamic, fiorini1998motion, wilkie2009generalized, van2011reciprocal, arul2021v, brito2019model, shen2020collision, singamaneni2021human, helbing1995social, ferrer2013robot, sebastian2017socially, boldrer2020socially}.
Supervised learning-based approaches are purely data-driven and easy to use, but require laboriously collecting a representative set of expert demonstrations to train the network \cite{pfeiffer2017perception, tai2016deep, loquercio2018dronet,  kahn2021badgr, pokle2019deep, xie2021towards}.
Reinforcement learning-based approaches are experience-driven and similar to human learning, but require carefully designing a reward function \cite{long2018towards, fan2020distributed, guldenring2020learning, perez2020robot, huang2021towards, chen2017socially, everett2018motion, everett2021collision, chen2019crowd, chen2020robot, liu2020robot, liu2020decentralized, chen2020relational, sathyamoorthy2020densecavoid, dugas2020navrep, xu2021human, pateldwa}.
Although each type of approach has its own advantages and disadvantages, we choose the reinforcement learning-based framework because it is difficult to manually design a general model or collect effective training data in uncontrolled and human-filled environments. 

With any of the different approaches to autonomous navigation, one key set of issues is which sensor data to use, how to process it, and which data representations to use.
Model-based approaches usually preprocess the raw sensor data and use the extracted high-level information, such as the obstacle costmaps \cite{fox1997dynamic, singamaneni2021human}, the relative velocity of obstacles \cite{fiorini1998motion, wilkie2009generalized, van2011reciprocal, arul2021v}, and the environment dynamics \cite{brito2019model, shen2020collision}.
On the other hand, end-to-end learning approaches focusing on static environments directly feed the raw sensor data, \eg lidars \cite{pfeiffer2017perception} or cameras \cite{tai2016deep, loquercio2018dronet, kahn2021badgr}, and goal information into CNNs.
DRL-based approaches are more varied, with some also feeding raw lidar scans (and goal points) into CNNs \cite{long2018towards, fan2020distributed, guldenring2020learning, perez2020robot, huang2021towards} while others extract and utilize the position and velocity of pedestrians \cite{chen2017socially, everett2018motion, everett2021collision}, model robot-human interactions \cite{chen2019crowd, chen2020robot, liu2020decentralized, liu2020robot}, predict pedestrian trajectories \cite{sathyamoorthy2020densecavoid, chen2020relational}, predict latent states \cite{dugas2020navrep}, or construct obstacle cost matrices \cite{pateldwa}.
The long-term success of model-based approaches and the results of the recent DRL studies demonstrate the utility of preprocessed data representations over raw sensor data, as they contain more useful information about pedestrian motion which, in turn, improves safety in dynamic environments.
Based on these trends, we will also utilize preprocessed sensor data, namely a short history of lidar data and the current kinematics of pedestrians, which we previously showed was able to improve navigation safety in a supervised setting \cite{xie2021towards}.
A significant advantage of this data representation is that it is much simpler and more easily interpretable than the complex pedestrian predictions or interactions used in other DRL-based works \cite{sathyamoorthy2020densecavoid, chen2020relational,chen2019crowd, chen2020robot, liu2020decentralized, liu2020robot}.
Also, all of the data is represented entirely within the robot's local coordinate frame, making the solution robust to errors in the localization, which can be common in densely crowded dynamic environments.

One of the biggest challenges in using DRL is designing an appropriate reward function. 
Most approaches include two terms, one to reward progress towards the goal and one to penalize collisions \cite{long2018towards, fan2020distributed, guldenring2020learning, perez2020robot, huang2021towards, chen2017socially, everett2018motion, everett2021collision, chen2019crowd, chen2020robot, liu2020robot, liu2020decentralized, chen2020relational, sathyamoorthy2020densecavoid, dugas2020navrep}. 
Although this simple reward function design can work well, their sparse collision avoidance reward function only gives a large negative reward to penalize collisions, and does not give a direct and effective reward signal to guide the robot to actively avoid obstacles. 
In fact, the passive collision avoidance reward function is a zero-order function, meaning it assumes the risk of collision depends only on the distance to the obstacle.
We argue that the \emph{relative motion} of pedestrians (and other objects) is more important since people often follow one another closely in dense crowds and give more space when walking in opposite directions.
One attempt at this is the reverse reward function from \cite{pateldwa}, which prioritizes safety but leads to slow motion and can cause deadlock in dense crowds.

We propose a novel reward function that uses first-order velocity obstacles to penalize headings that are likely to lead to collisions, thereby guiding the robot to continuously tune its heading direction to actively avoid crowds of pedestrians while making progress towards the goal.
Note that Han et al.~\cite{han2022reinforcement} simultaneously developed a reciprocal velocity obstacle (RVO)-based reward function for distributed multi-robot navigation.
This is similar to our velocity obstacle concept but has two main differences.
First, our work and theirs solve two distinct, albeit related, problems.
We seek to prevent \emph{robot-human} collisions by using VOs, which place the burden of collision avoidance entirely on the robot as we cannot directly model or control human behavior.
On the other hand, Han et al.~\cite{han2022reinforcement} seek to prevent \emph{robot-robot} collisions by using RVOs, which split the burden of collision avoidance between the agents based on the assumption that both agents run similar control policies.
Second, our approach encourages the robot to actively steer towards the goal while avoiding potential future human-robot collisions while theirs encourages the robot to focus only on potential robot-robot collisions, as judged by the expected collision time, and ignore the goals.
Sun et al.~\cite{sun2019crowd} also use RVOs in tandem with a DRL-based control policy for multi-robot navigation.
However, instead of using RVOs in the DRL reward function, they use it as a parallel control policy, switching from DRL to RVO when a collision is imminent.

Inspired by the problems and approaches mentioned above, this paper starts with two core issues of sensor data representation and reward function design to solve the problem of autonomous navigation in dynamic crowds.
We use a unique combination of preprocessed sensor data as the input to our control policy. 
Specifically, we have information about static obstacles and the geometric layout of the environment from a short history of lidar data, kinematic information about dynamic obstacles from a multi-object tracker, and information about how to reach the destination from a sub-goal point.
In designing our novel reward function, we leverage the utility of model-based approaches (\ie velocity obstacles) to create a navigation policy that does not have the same drawbacks as velocity obstacles.
Specifically, our policy does not require the robot to have perfect information about pedestrian kinematics.
This new term will allow our robot to learn a robust navigation policy with a good balance between collision avoidance and speed.

\section{Navigation Policy}
\label{sec:safe_and_fast_navigation_policy}
In this section, we formulate the navigation problem, with a focus on safety and speed through dynamic environments.
We then detail our DRL-based approach, describing the preprocessed observation space, the DRL network architecture, and the novel VO-based reward function design.

\begin{figure*}[t]
    \centering
    \includegraphics[width=0.98\textwidth]{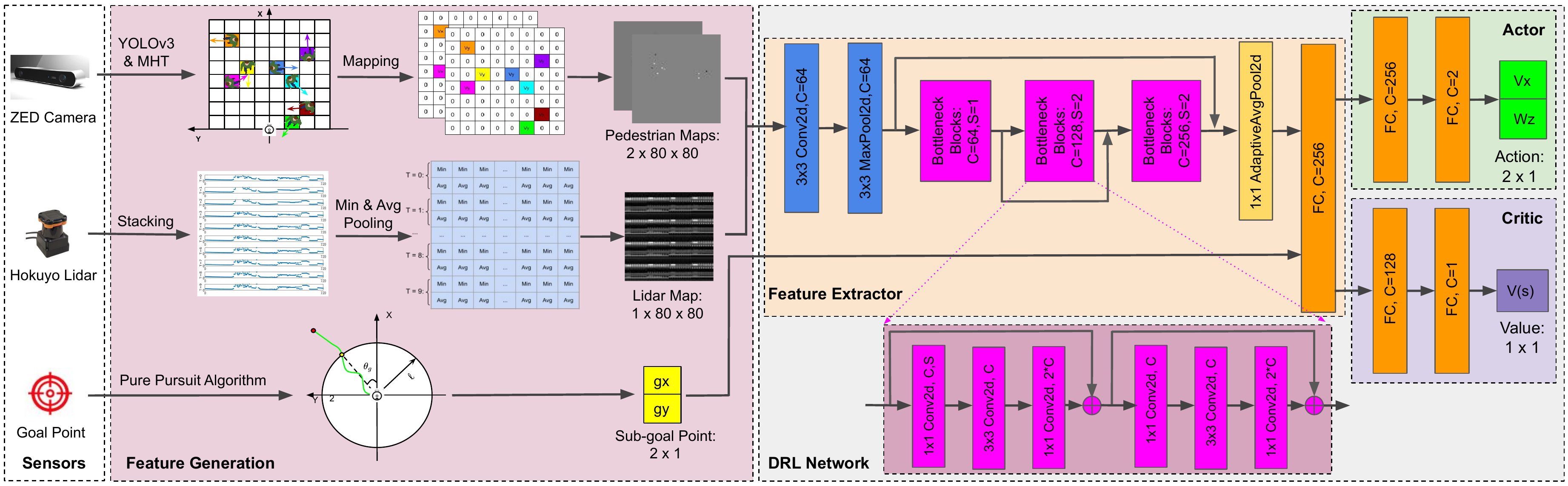}
    \caption{
    Overall system architecture of our DRL-VO control policy. 
    Raw sensor data from the ZED camera and Hokuyo lidar, as well as the goal point, are fed into a feature generation module to create intermediate features.
    Three $80 \times 80$ intermediate-level feature maps, one from lidar and two containing pedestrian information, along with the relative position of the sub-goal are fed to the feature extractor network, and the resulting high-level features are fed to the actor and critic networks separately.
    The critic network outputs the state value and the actor network outputs the linear and angular steering velocities.
    The feature extractor module consists of bottleneck residual blocks \cite{he2016deep}, while the critic and actor modules consist of fully connected layers. 
    ``$3\times3$'' or ``$1\times1$'' denote the kernel size, and ``C'' and ``S'' denote the number of output channels and the stride length, respectively.
    }
    \label{fig:cnn}
\end{figure*}

\subsection{Problem Formulation}
\label{subsec:problem_formulation}
Due to the limited field of view (FOV) of sensors (\eg lidars and cameras), there is no complete environmental perception for the robot in the real world.
Extracting and processing useful information from sensors to obtain the partial observation of the environment $\mathbf{o^t}$ is the first step for the robot to autonomously and quickly navigate through complex and human-filled environments. 
The robot then feeds this partial observation $\mathbf{o^t}$ into a control policy $\mathbf{\pi_{\mathbf{\theta}}}$ to compute the suitable steering action $\mathbf{a^t}$, which is defined as 
\begin{equation}
    \mathbf{a^t} \sim \mathbf{\pi_{\mathbf{\theta}}}(\mathbf{a^t}|\mathbf{o^t}),
    \label{eq:pm}
\end{equation}
where $\mathbf{\theta}$ are our control policy parameters.
This complicated navigation decision-making process can be formulated as a partially observable Markov decision process (POMDP), which is denoted by a 6-tuple $\left \langle S,A,T,R,\Omega,O \right \rangle$ where $S$ is the state space, $A$ is the action space, $T$ is the state-transition model, $R$ is the reward function, $\Omega$ is the observation space, and $O$ is the observation probability distribution. 

One of the most well-known tools for solving large POMDPs is deep reinforcement learning \cite{gosavi2009reinforcement}.
The general objective $L\left(\theta\right)$ of DRL is to maximize the expected discounted return by optimizing the neural network-based policy $\mathbf{\pi_{\mathbf{\theta}}}$:
\begin{equation}
    L \left( \theta \right) = \mathbb{E}_{\mathbf{\pi_{\mathbf{\theta}}}} \left[ \sum^\infty_{t=0} \gamma^t R^t \right],
    \label{eq:rl}
\end{equation}
where $\gamma$ is the discount rate in the range of $[0,1]$, $\mathbb{E}_{\mathbf{\pi_{\mathbf{\theta}}}}\left[ \cdot \right]$ denotes the expected value of a random variable assuming the agent follows the neural network-based policy $\mathbf{\pi_{\mathbf{\theta}}}$.
Figure~\ref{fig:cnn} outlines our DRL network, which consists of four modules: feature generation, feature extractor, actor, and critic.
The feature generation module generates our intermediate features, \ie the partial observation $\mathbf{o}^t$, from the raw sensor data.
The feature extractor module then extracts high-level features from $\mathbf{o}^t$, the critic module generates the state value, and the actor module generates steering action $\mathbf{a}^t$.

\subsection{Observation Space}
\label{subsec:obervation_space}
We use the same partial observation $\mathbf{o}^t = [\mathbf{l}^t, \mathbf{p}^t, \mathbf{g}^t]$ as our previous work using supervised learning \cite{xie2021towards}, which consists of three preprocessed components: lidar history ($\mathbf{l}^t$), pedestrian kinematics ($\mathbf{p}^t$), and the sub-goal position ($\mathbf{g}^t$).
All the data in the observation $\mathbf{o}^t$ are expressed in the local reference frame of the robot.
This allows our method to be robust to errors in robot localization with respect to a global reference frame, which happens more frequently in densely-packed dynamic environments as there are fewer stationary landmarks for the robot to localize itself against. 
Additionally, relative data is more natural for planning purposes since navigating in a dense crowd is more about going with the flow of traffic than meeting some absolute velocity constraints.

We will briefly summarize the process for extracting the observation data $\mathbf{o}^t$ from the raw sensor data below, with the full details available in \cite{xie2021towards}. 
One of the key features of this previous work was the use of handcrafted intermediate features which allowed us to set the size of the control policy.
We do this by using a consistent data format for all input types, which is an $80 \times 80$ grid in our case.

\subsubsection{Pedestrian Kinematics}
\label{subsubsec:pedestrian_kinematics}
Unlike complex CAGE representations for crowd flow prediction \cite{sohn2020laying}, which only encode crowd location information, our proposed simple pedestrian kinematic map representation encodes both the location and velocity information of detected pedestrians.
Specifically, we encode the pedestrian kinematics in a pair of $80 \times 80$ grids that contain information about both the relative locations and velocities of all pedestrians in a \unit[$20\times20$]{m} area in front of the robot. 
To create these grids, we first feed the RGB image and 3-D point cloud coming from a depth camera (\ie ZED camera) into the YOLOv3 \cite{redmon2018yolov3} detector to detect pedestrians and extract their relative locations.  
Once we extract these instantaneous estimates, we use a multiple hypothesis tracker (MHT) \cite{yoon2018multiple} to track the \emph{relative} positions and velocities of pedestrians.
The velocity values are stored in the grid, using the position values to determine the cell index, as the ZED Camera track in the Feature Generation block of Fig.~\ref{fig:cnn} shows.

\subsubsection{Lidar History}
\label{subsubsec:lidar_history}
The lidar data $\mathbf{l}^t$ consists of a history of \unit[0.5]{s} data (\ie 10 scans), where we apply a combination of minimum and average pooling to each scan. The resulting pooled data is stacked \unit[4]{times} to create an $80 \times 80$ array, as the Hokuyo Lidar track in the Feature Generation block of Fig.~\ref{fig:cnn} shows. This data format is identical to our previous work \cite{xie2021towards}, but here we present the results of ablation studies used to select this data format and the corresponding parameters, such as the history length.

\begin{figure}[t]
    \centering
    \subfigure[Lidar preprocessing steps]
    {
        \centering
        \includegraphics[width=0.4\textwidth]{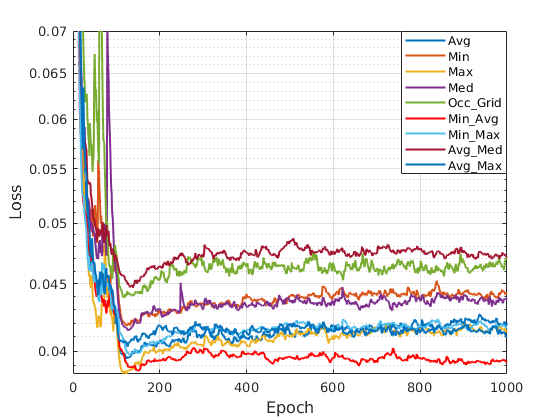}
        \label{fig:lidar_preprocessing_loss}
    } 
    \subfigure[Lidar history length]
    {
        \centering
        \includegraphics[width=0.4\textwidth]{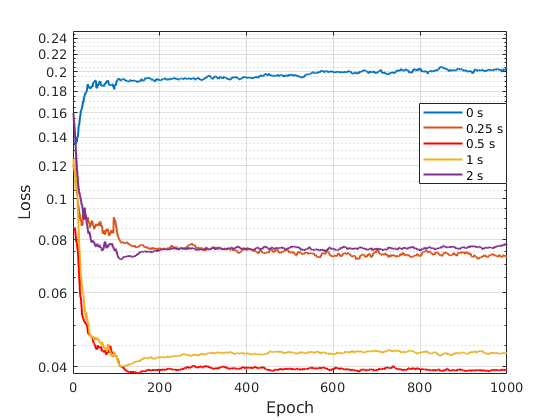}
        \label{fig:lidar_time_window_loss}
    }%
    \caption{Validation loss curves of different lidar data configurations. The loss is the MSE between the expert velocity and the learned velocity of the resulting control policy.}
    \label{fig:lidar_loss}
\end{figure}

First, motivated by the lidar data downsampling operation from \cite{pfeiffer2018reinforced}, we explore the use of different lidar preprocessing operations, including reshaping the raw lidar data, projecting scan points into an occupancy grid map, and pooling (minimum, average, median, and maximum).
We use each of these different operations, and several combinations thereof, to train our navigation policy in a supervised manner using the dataset from \cite{xie2021towards}.
Figure~\ref{fig:lidar_preprocessing_loss} shows the resulting mean square error (MSE), \ie the loss, between the learned and expert velocities for each resulting control policy.
We found that using a combination of minimum pooling and average pooling to extract two separate distance measurements per scan region, as Fig.~\ref{fig:lidar_preprocessing_loss} shows, yielded the best performance.

Once we determined the most informative way to structure the lidar data, we then studied how long of a history of scans to keep.
We considered time windows of 0, 0.25, 0.5, 1, and \unit[2]{seconds},\footnote{Note: we sample the lidar data at \unit[20]{Hz}, so each time corresponds to a different number of scans.} using the same training and validation procedure as we did when studying the different lidar preprocessing operations.
Figure~\ref{fig:lidar_time_window_loss} shows the resulting MSE loss curve for each resulting control policy.
We see that a short history of lidar data contains more useful information than only the current scan (\ie \unit[0]{s}), while keeping data from beyond a certain time period resulted in no improvement or even degraded performance.
We found the best time window to be \unit[0.5]{s} (\ie 10 lidar scans), which we conceptually think of as the effective time constant for pedestrian information.

\subsubsection{Goal Position}
\label{subsubsec:sub-goal_position}
Lastly, the robot needs to know the relative position of the goal.
Instead of feeding the final goal point to our DRL network, we use the sub-goal point along a nominal path to the final goal as this allows the robot to traverse long paths through complex, non-convex environments and avoid issues with normalization as the distance to the sub-goal is constant.
To select the sub-goal $\mathbf{g^t}$, we use the pure pursuit algorithm \cite{coulter1992implementation}, which calculates the point along the full path that is a specified distance (\unit[2]{m} in our case) ahead of the robot. 
By tuning this distance, the robot can look far ahead or right nearby.
Figure~\ref{fig:cnn} illustrates this process in the Goal Point track of the Feature Generation block.
We hypothesize that this sub-goal information will enable the robot to more accurately navigate to the final goal position.

\subsubsection{Normalization}
We use the Max-Abs scaling procedure to normalize all observation data to $[-1, 1]$ before feeding them into our control policy network:
\begin{equation}
    \mathbf{o}^t= 2 \frac{\mathbf{o}^t - \mathbf{o}^t_{\rm min}}{\mathbf{o}^t_{\rm max} -\mathbf{o}^t_{\rm min}} - 1,
    \label{eq:norm_p}
\end{equation}
where $\mathbf{o}^t_{\rm max}$ and $\mathbf{o}^t_{\rm min}$ are the maximum value and minimum value of the data (\eg for the lidar history $\mathbf{l}^t$, they are the minimum and maximum range value of the lidar sensor).

\subsection{Action Space}
\label{subsec:action_space}
The action $\mathbf{a}^t=[v^t_x, \omega^t_z]$ of our DRL control policy has two terms, the forward ($v^t_x$) and rotational ($\omega^t_z$) velocities in the local robot frame. 
Note, we use a continuous action space as we believe this gives the robot more accurate control.
We limit the range of the forward velocity $v^t_x$ to \unit[{$[0,0.5]$}]{m/s} and the rotational velocity $\omega^t_z$ to \unit[{$[-2,2]$}]{rad/s} based on the limitations of the Turtlebot2, our target hardware platform.
Note that similar to observation space, we also normalize the action space to [-1, 1] using the Max-Abs scaling procedure.

\subsection{Network Architecture} 
\label{subsec:network_architecture}
The feature extractor network is identical to the backbone CNN network of our previous work \cite{xie2021towards}, which fuses the lidar historical observation $\mathbf{l}^t$ and pedestrian kinematics observation $\mathbf{p}^t$.
After extracting high-level features, a single fully connected layer is used to fuse those high-level features with the sub-goal information.
Both the actor module and the critic module are simply constructed from two fully connected layers.
Note that each convolutional layer in our feature extractor module is followed by a batch normalization layer and an rectified linear
unit (ReLU) action layer.
Additionally, the last fully connected layer in our feature extractor is followed by a ReLU action layer while the last fully connected layers of the actor and critic modules have no ReLU action layers.

Our architecture is an example of a middle fusion network, as data from different input streams are fused in the middle of the chain between raw data input and output.
Feng et al.~\cite{feng2020deep} provide a detailed discussion about different fusion architectures (in a general setting), which are typically categorized as early, middle, or late fusion.
They note that there is no solid evidence to support which fusion network is the best choice, though there is a connection between the computational load of a neural network and its structure as the number of layers and their sizes affect both the number of computations that must be done to produce and output and the memory required to store the parameters of the neural network.
To examine this, we used the same dataset and supervised training manner as our previous work~\cite{xie2021towards} to train two networks.
The first, which we call Middle, fuses the pedestrian kinematics and lidar data at the input to the feature extractor block, as we do above, and the second, called Late, fuses all data just before the output of the feature extractor block.
As we can see in Table~\ref{tab:fusion}, the Early network is about half the size of the Late network structure, which nearly doubles the processing rate and yields slightly better regression performance, with smaller root mean square error (RMSE) and explained variance ratio (EVA).

\begin{table}[t]
    \small\sf\centering
    \caption{Ablation experiment results with different fusion structures}
    \scalebox{0.92}{
        \begin{tabular}{c c c c c}
            \toprule
            \textbf{Network Structures} 
            & \textbf{RMSE} 
            & \textbf{EVA} 
            & \textbf{\# of Params} & 
            \textbf{FPS}\\
            \midrule
            Middle fusion  & \textbf{0.17} & \textbf{0.15} & \textbf{28.94 M} & \textbf{255.13} \\
            \hline
            Late fusion  & 0.18 & 0.17 & 57.86 M & 140.53 \\
           \bottomrule
        \end{tabular}
    }
    \label{tab:fusion}
\end{table}

Like our network, most other neural network-based control policies \cite{pokle2019deep, huang2021towards, sathyamoorthy2020densecavoid, perez2020robot} also use late or middle fusion architectures as this allows for flexible design because the different input sources each yield some learned intermediate representations which are all combined downstream.
The key difference between these and our architecture is that we use handcrafted features as our intermediate representation, much like a traditional model-based algorithm, rather than learned features, as are typical in learning-based algorithms.
One benefit of this approach is that our data representations, \ie the pedestrian map, are independent of the specific object detection and tracking tools used.
This allows us to easily swap out YOLO and the MHT for different approaches with no need for retraining.

\subsection{Reward Function}
\label{subsec:reward_function}
An essential problem of reinforcement learning methods is how to design a good reward function to guide the agent to learn desired behaviors.
Navigation tasks have two competing objectives, we want the robot to move as quickly as possible to reach the goal in minimum time but also safely to avoid colliding with any stationary or moving objects.
Thus, we design a multi-objective reward function:
\begin{equation}
    r^t = r^t_g + r^t_c + r^t_w + r^t_d,
    \label{eq:reward}
\end{equation}
where $r^t_g$ rewards making progress towards the goal, $r^t_c$ penalizes passively approaching or colliding with an obstacle, $r^t_w$ penalizes rapid changes in direction, and $r^t_d$ rewards actively steering to avoid obstacles and point towards the sub-goal.

Note that most previous works \cite{long2018towards, fan2020distributed, guldenring2020learning, perez2020robot, huang2021towards, chen2017socially, everett2018motion, everett2021collision, chen2019crowd, chen2020robot, liu2020robot, liu2020decentralized, chen2020relational, sathyamoorthy2020densecavoid, dugas2020navrep} only use such a zero-order passive collision avoidance reward $r^t_c$, a position-based heuristics, to guide robots to passively avoid obstacles.

While these standard terms are useful to learn good navigation behavior, we want to allow robots to indirectly learn from successful model-based navigation policies.
To do this, we introduce a new reward term $r^t_d$ that uses velocity obstacles \cite{fiorini1998motion} to set a desired navigation direction.
This term allows the robot to use first-order kinematic information (\ie velocity) to proactively avoid collisions while making progress towards the goal.

\subsubsection{Reaching the Goal}
\label{subsubsec:reaching_the_goal}
The reward is given by
\begin{equation}
    r^t_g =
    \begin{cases}
    r_{\rm goal} & \text{if $\|p_g^t\|) < g_m$} \\
    -r_{\rm goal} & \text{else if $t \geq t_{\rm max}$} \\
    r_{\rm path}(\|p_g^{t-1}\| - \|p_g^t\|) &\text{otherwise},
    \end{cases}
    \label{eq:r_g}
\end{equation}
where $p_g^t$ is the goal position (in the robot's frame) at time $t$.

We use $r_{\rm goal} = 20$, $r_{\rm path} = 3.2$, $g_m = \unit[0.3]{m}$, and $t_{\rm max} = \unit[25]{s}$.

\subsubsection{Passive Collision Avoidance}
\label{subsubsec:passive_collision_avoidance}
The reward is given by
\begin{equation}
    r^t_c =
    \begin{cases}
    r_{\rm collision} & \text{if $\|p^t_o\| \leq d_r$} \\
    r_{\rm obstacle}(d_m - \|p^t_o\|) &\text{else if $\|p^t_o\| \leq d_m$} \\
    0 &\text{otherwise},
    \end{cases}
    \label{eq:r_c}
\end{equation}
where $p^t_o$ is obstacle position at time $t$. We use $r_{\rm collision} = -20$, $r_{\rm obstacle} = -0.2$, $d_r = \unit[0.3]{m}$, and $d_m = \unit[1.2]{m}$.

\subsubsection{Path Smoothness}
\label{subsubsec:path_smoothness}
The reward is given by 
\begin{equation}
    r^t_w =
    \begin{cases}
    r_{\rm rotation}|\omega^t_z| & \text{if $|\omega^t_z| > \omega_m$} \\
    0 &\text{otherwise},
    \end{cases}
    \label{eq:r_w}
\end{equation}
where $r_{\rm rotation} = -0.1$ and $\omega_m = \unit[1]{rad/s}$.

\begin{figure}[t]
    \centering
    \includegraphics[width=0.45\textwidth]{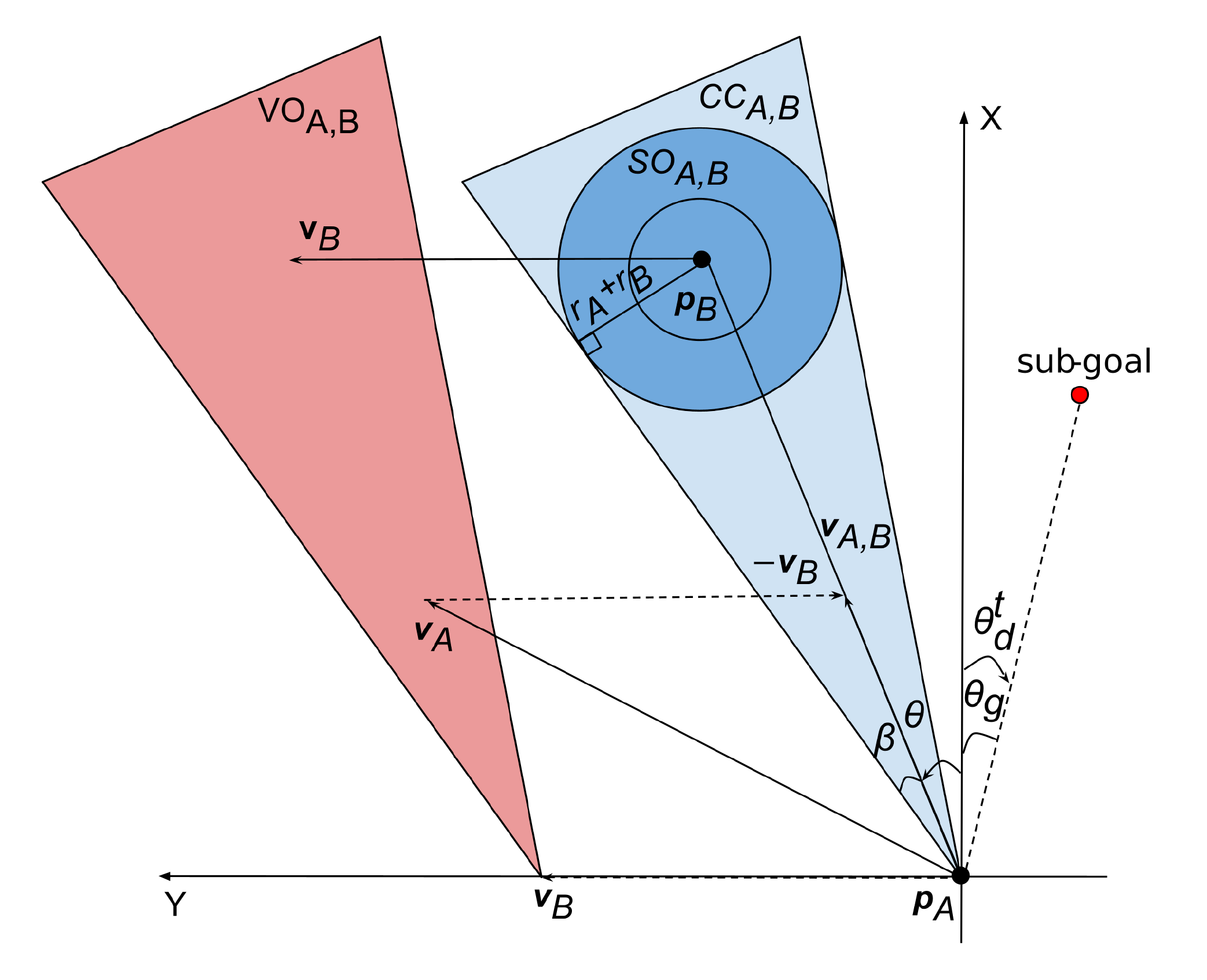}
    \caption{The pictogram for velocity obstacle $VO_{A,B}$ (red cone),  collision cone $CC_{A,B}$ (light blue cone), and special occupancy $SO_{A,B}$ (blue circle).
    $A$ represents the robot with radius $r_A$, position $\mathbf{p}_A$, and velocities $\mathbf{v}_A$.
    $B$ represents the moving pedestrian with radius $r_B$, position $\mathbf{p}_B$, and velocities $\mathbf{v}_B$.
    $\mathbf{v}_{A,B} = \mathbf{v}_A-\mathbf{v}_B$ represents the relative velocity of $A$ with respect to $B$.
    Note that all quantities are in the robot's local frame.
    }
    \label{fig:velocity_obstacle}
\end{figure}

\subsubsection{Active Heading Direction}
\label{subsubsec:active_heading_direction}
The reward is given by
\begin{equation}
    r^t_d =
    r_{\rm angle}(\theta_m - |\theta^t_d|),
    \label{eq:r_d}
\end{equation}
where $\theta^t_d$ is the desired heading direction in the robot's local frame and $\theta_m$ is the maximum allowable deviation of the heading direction. We use $r_{\rm angle} = 0.6$ and $\theta_m = \unit[\frac{\pi}{6}]{rad}$.

The key to this reward term is to find the desired direction $\theta^t_d$ that moves towards the goal while also being collision-free.
To do this, we extend the concept of velocity obstacles \cite{fiorini1998motion}.
The velocity obstacle $VO_{A,B}$ is the velocity space where the current velocity of robot $A$ would cause a collision with an obstacle $B$ at some future time, as Fig.~\ref{fig:velocity_obstacle} shows.
To define the velocity obstacle $VO_{A,B}$, we first need to define the special occupancy region:
\begin{equation}
    SO_{A,B} =
     \left\{ \mathbf{p}_S \mid d(\mathbf{p}_S,\mathbf{p}_B) < r_A + r_B \right\},
    \label{eq:so}
\end{equation}
where $d(\mathbf{p}_S, \mathbf{p}_B)$ is the distance between $\mathbf{p}_S$ and $\mathbf{p}_B$.
Then, the collision cone is defined as
\begin{equation}
    CC_{A,B} =
    \left\{ \mathbf{v}_{A,B} \mid \exists t,  \mathbf{v}_{A,B} t \cap SO_{A,B} \neq \emptyset \right\}.
    \label{eq:cc}
\end{equation}
Finally, the velocity obstacle $VO_{A,B}$ is defined as
\begin{equation}
    VO_{A,B} = CC_{A,B} \oplus \mathbf{v}_B,
    \label{eq:vo}
\end{equation}
where $\oplus$ denotes the Minkowski vector sum operator.

The physical meaning of $CC_{A,B}$ is that any relative velocities $\mathbf{v}_{A,B} \in CC_{A,B}$ will cause a collision at a future time.
From the perspective of the direction angle of $\mathbf{v}_{A,B}$, the collision cone $CC_{A,B}$ can also be defined as
\begin{equation}
    CC_{A,B} \in \left[ \theta - \beta, \theta + \beta \right]
    \label{eq:cc_d},
\end{equation}
where $\theta$ and $\beta$ can be easily calculated from Fig.~\ref{fig:velocity_obstacle} using simple geometric relationships:  
\begin{align}
    \theta &= \arctan2\left(\frac{\mathbf{p}_{B_y}}{\mathbf{p}_{B_x}}\right),
    &
    \sin{\beta} &= \frac{r_A+r_B}{\|\mathbf{p}_B\|}.
    \label{eq:thetabeta}
\end{align}

\begin{algorithm}[t]
    \DontPrintSemicolon
    \caption{Search desired direction angle}
    \label{alg:search_direction}
    \KwInput {Sub-goal direction angle $\theta_{g}$, pedestrians from MHT $B_{\rm peds}$, robot linear velocity $\mathbf{v}_{A_x}$, number of samples $N$}
    \KwOutput {Optimal direction angle $\theta^t_d$}
    {\textbf{initialize:} $\theta^t_d \gets \frac{\pi}{2}$} \;
    \If {$B_{\rm peds} \neq \emptyset$}
    {
        $\theta_{\rm min} \gets \infty$ \;
        \For {$i = 1,2,\ldots,N$}
        {
            $\theta_u$ $\gets$ sample from $[-\pi, \pi]$ \;
	        $\rm free$ $\gets$ True \;
	        \For {$B$ in $B_{\rm peds}$}
	        {
        	        $\theta_{v_{A,B}} \gets \atan2\left(\frac{\mathbf{v}_{A_x}\sin(\theta_u) - \mathbf{v}_{B_y}}{\mathbf{v}_{A_x}\cos(\theta_u) - \mathbf{v}_{B_x}} \right)$ \;
        	        $\theta, \beta$ $\gets$ from \eqref{eq:thetabeta} using $B$\;
        	        \If {$\theta_{v_{A,B}} \in [\theta - \beta, \theta + \beta]$}
        	        {
        	            $\rm free$ $\gets$ False \;
        	            \textbf{break} \;
        	        }
        	}
        	\If {$\rm free$}
        	{
        	    \If {$ \lVert \theta_u - \theta_{g} \rVert < \theta_{\rm min}$}
        	    {
        	        $\theta_{\rm min} \gets \lVert \theta_u - \theta_{g} \rVert$ \;
        	        $\theta^t_d \gets \theta_{u}$ \;
        	    }
        	}
        }
    }
    \Else 
    {
        $\theta^t_d \gets \theta_{g}$ \;
    }
    \Return $\theta^t_d$
\end{algorithm}

Using the collision cone $CC_{A,B}$, the robot can know which heading direction angles will cause collisions with all of the moving pedestrians tracked by MHT.
The robot then uses these collision cones and the direction of the sub-goal ($\theta_{g}$) to find the desired heading direction angle $\theta^t_d$ using Algorithm~\ref{alg:search_direction}. 
This sampling-based search algorithm is motivated by \cite{wilkie2009generalized}.

\subsection{Deep Reinforcement Learning Algorithm}
\label{subsec:drl_algorithm}
While there are many deep reinforcement learning algorithms, we use the proximal policy optimization (PPO) algorithm \cite{schulman2017proximal} to train our DRL network.
A major advantage of the PPO algorithm is that it is much simpler to implement and can achieve comparable or better results than other state-of-the-art algorithms (\eg A2C~\cite{mnih2016asynchronous}, TRPO~\cite{schulman2015trust} and ACER~\cite{wang2016sample}).
In addition, many other robot navigation works~\cite{long2018towards, fan2020distributed, guldenring2020learning, pateldwa} use PPO and have found that it works well on navigation problems. 
As suggested by \cite{schulman2017proximal}, we use Adam optimizer \cite{kingma2014adam}, a stochastic gradient descent method, to find the optimal policy parameters $\theta^\ast$.

\begin{figure}[t]
    \centering
    \subfigure[Lobby]{
            \centering
            \includegraphics[width=0.486\linewidth]{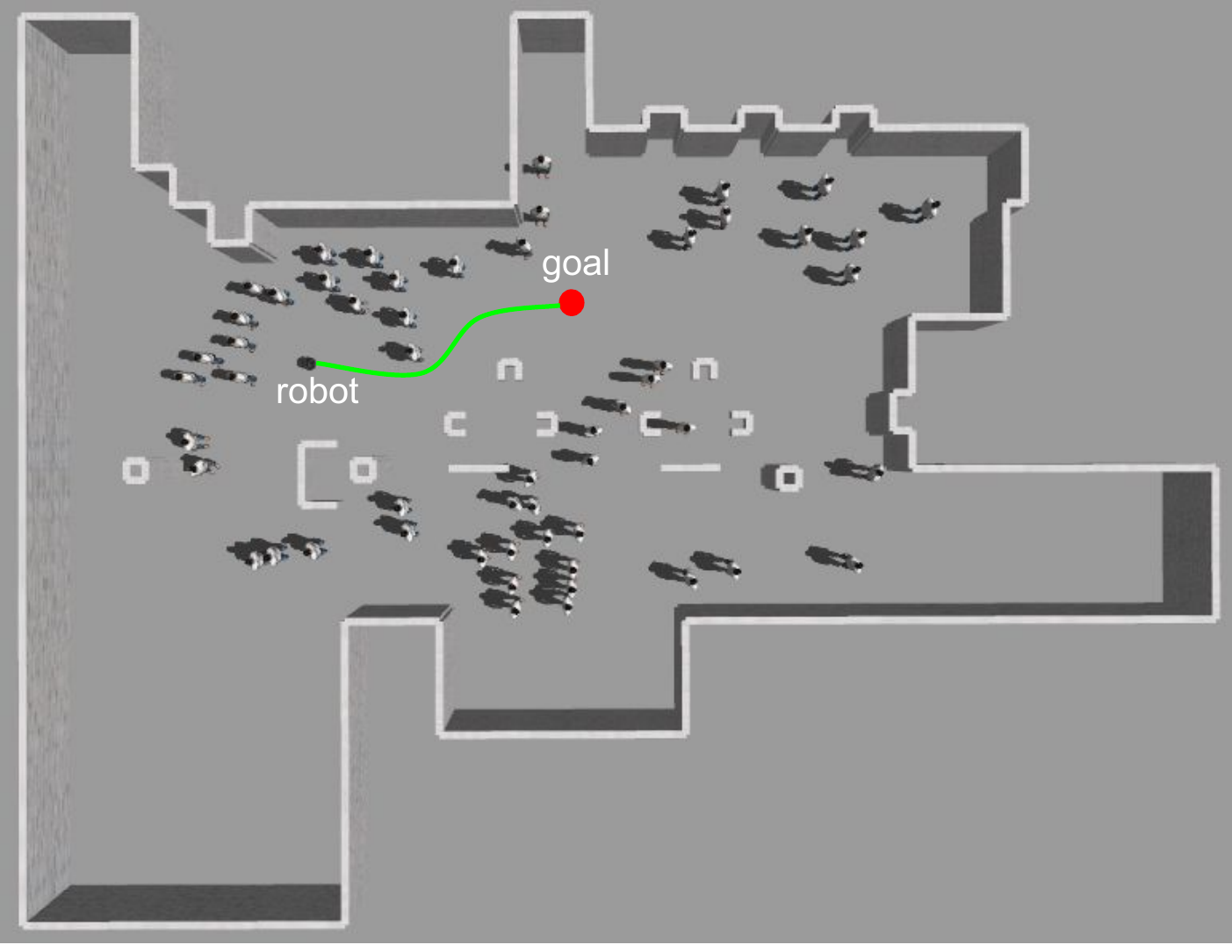}
            \label{fig:lobby}
    }%
    \subfigure[Autolab]{
            \centering
            \includegraphics[width=0.45\linewidth]{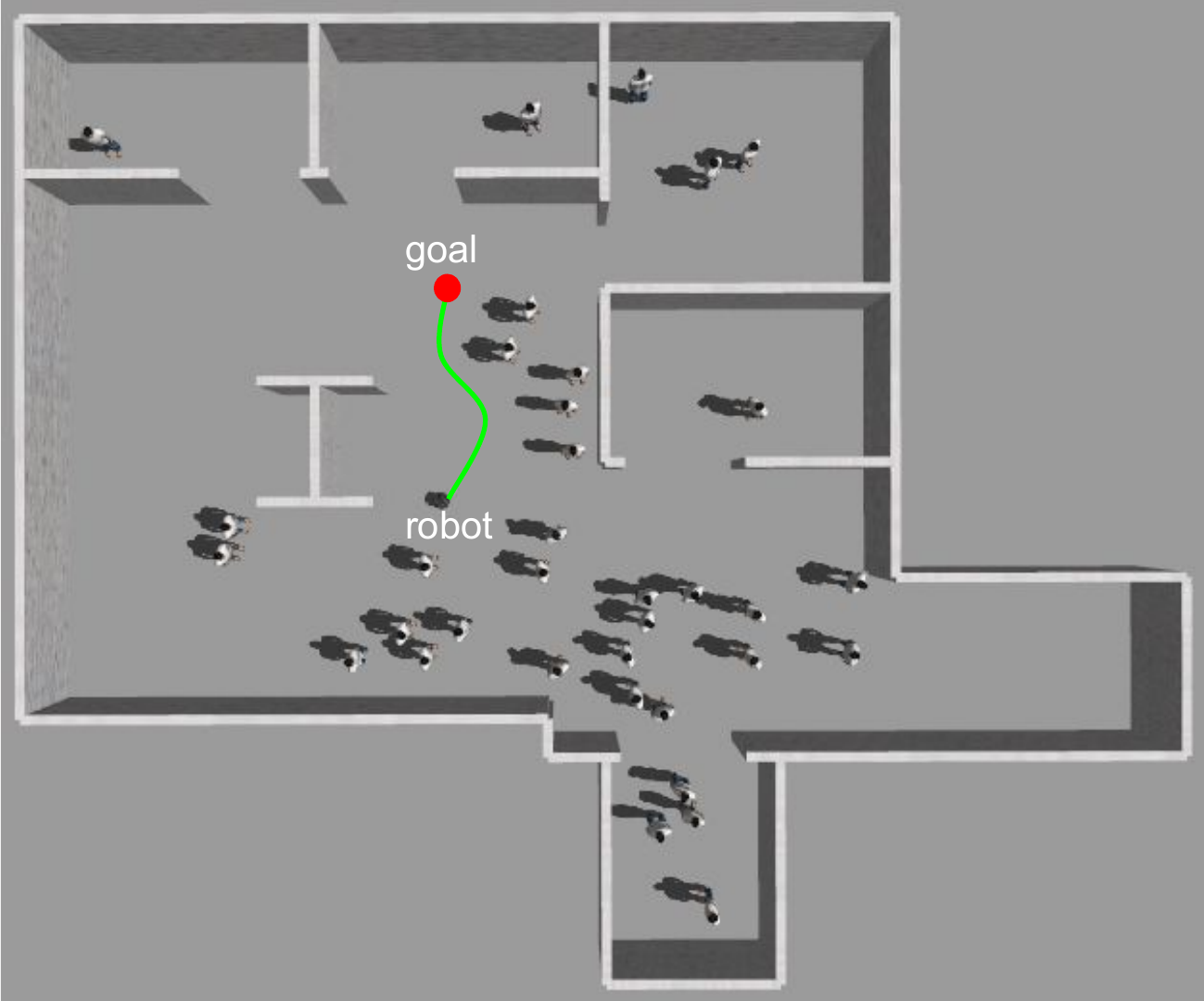}
            \label{fig:autolab}
    }%
    \vspace{0.01cm}
    \subfigure[Cumberland]{
            \centering
            \includegraphics[width=0.45\linewidth]{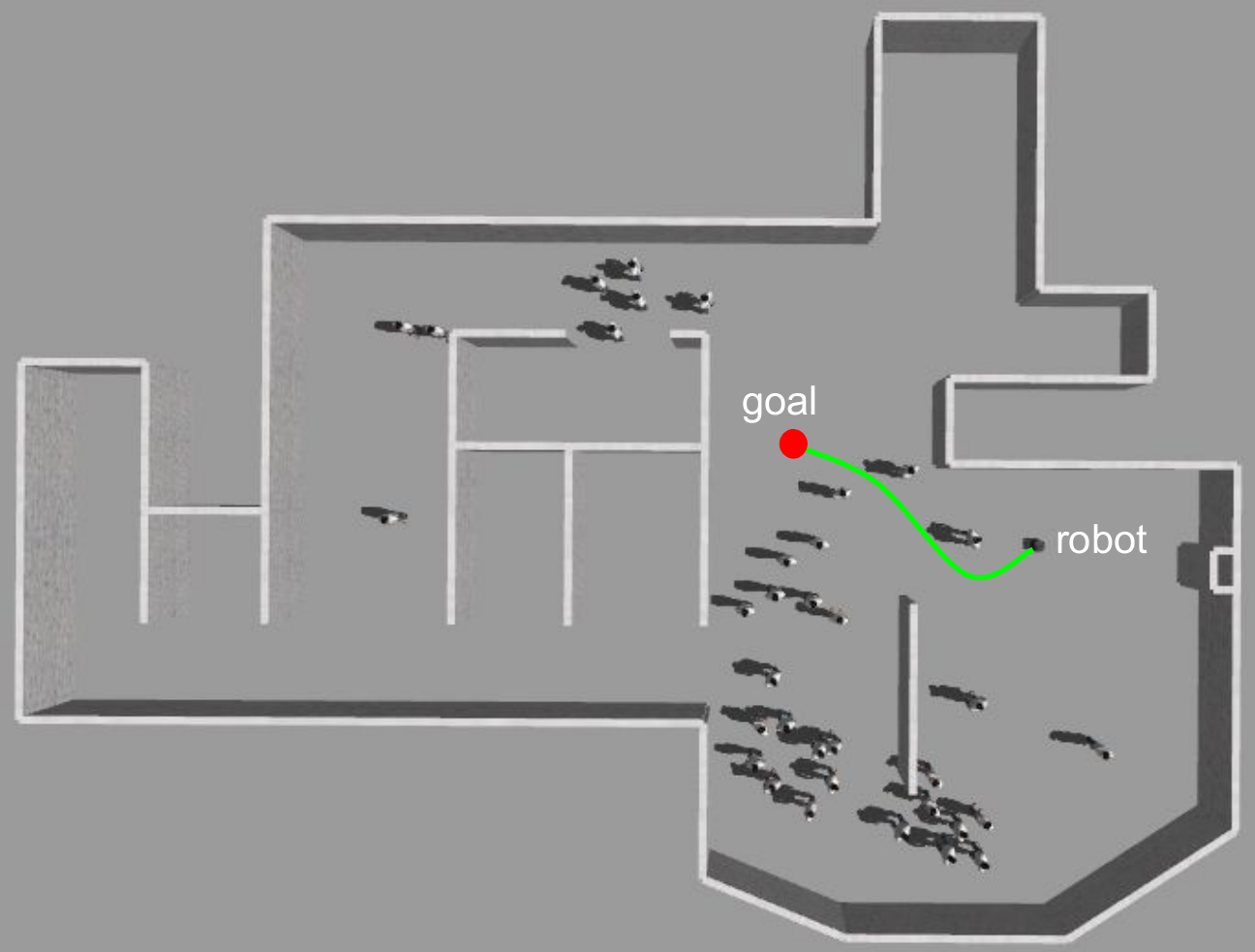}
            \label{fig:cumberland}
    }%
    \subfigure[Freiburg]{
            \centering
            \includegraphics[width=0.484\linewidth]{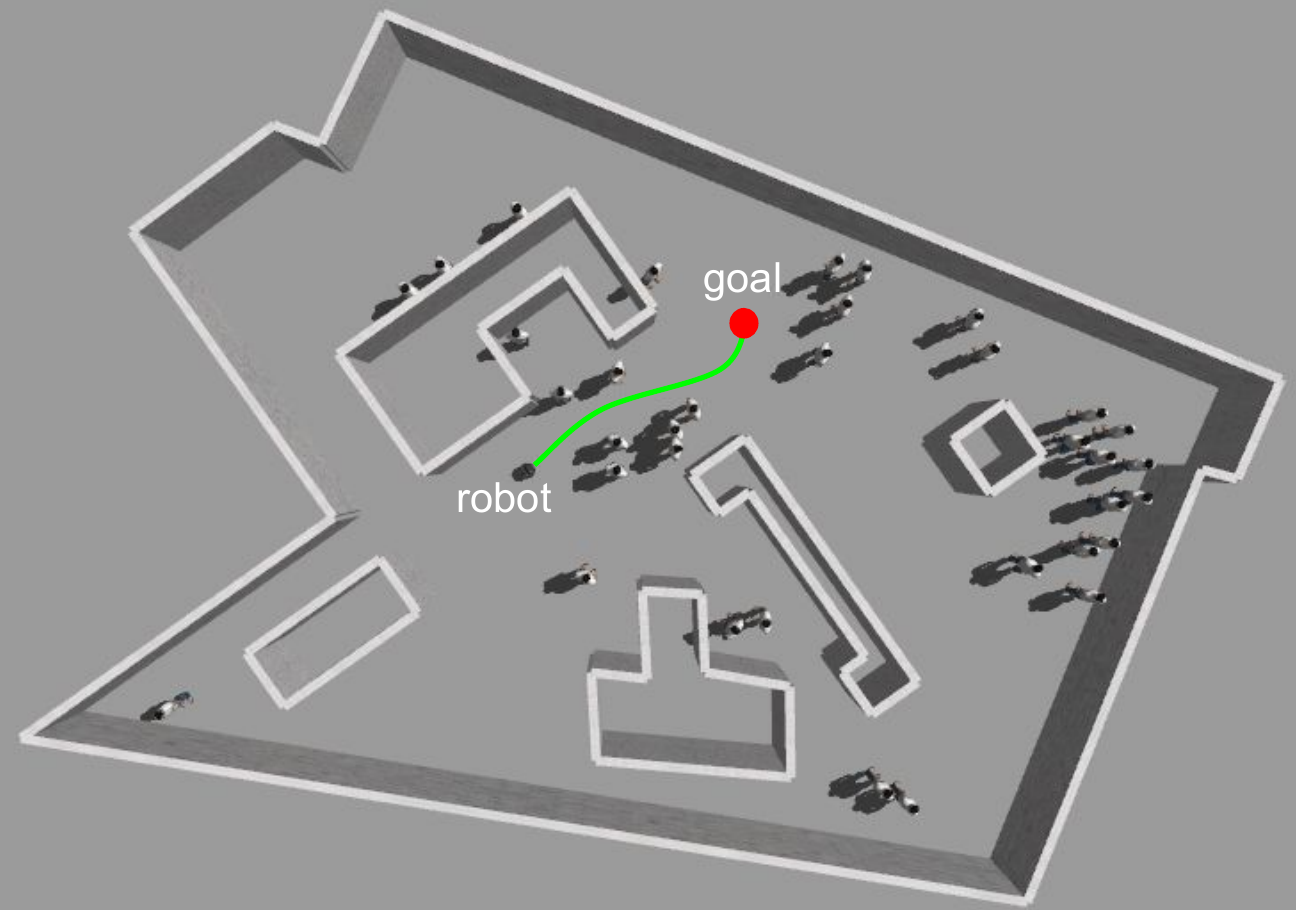}
            \label{fig:frieburg}
    }%
    \vspace{0.01cm}
    \subfigure[Square]{
            \centering
            \includegraphics[width=0.43\linewidth]{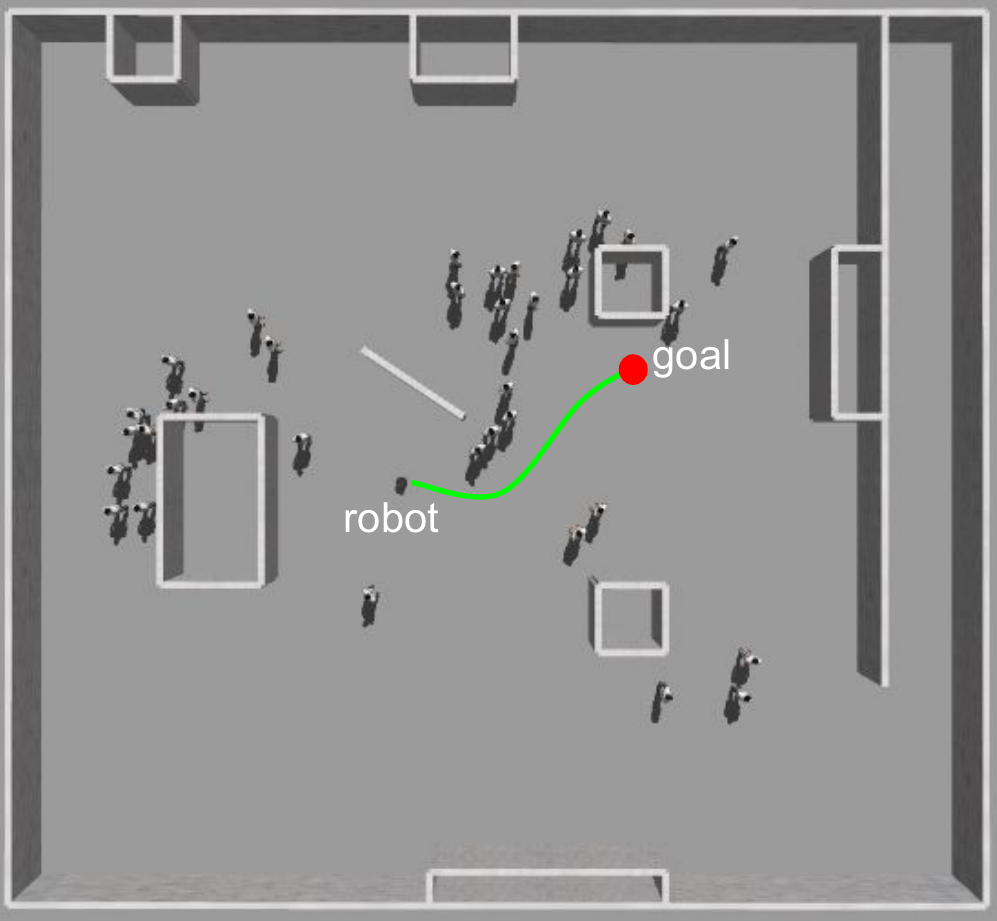}
            \label{fig:square}
    }%
    \caption{Gazebo simulation environments. 
             The Lobby world has two configurations with 34 pedestrians and 5-55 (sampling interval 10) pedestrians respectively.
             The Autolab world, Cumberland world, Freiburg world, and Square world only have one configuration with 25-35 (sampling interval 10) pedestrians.
    }
    \label{fig:gazebo}
\end{figure}

\section{Crowded Dynamic Environments}
\label{sec:dynamic_results}
To demonstrate the efficacy and performance of our proposed control policy, we first use a Gazebo simulator \cite{koenig2004design} and the PEDSIM library \cite{gloor2016pedsim} to conduct a set of 3D simulated experiments, and then use a Turtlebot2 robot to conduct the real-world experiments.
This section describes the experimental setup, training procedure, and results.

\subsection{Experimental Setup}
\label{subsec:experimental_setup}
\subsubsection{Robot Configuration}
For the simulated and hardware tests, we use a Turtlebot2 robot with a ZED stereo camera and a Hokuyo UTM-30LX lidar, as shown in Fig.~\ref{fig:navigation_examples}.
The maximum velocity of the Turtlebot2 is limited to \unit[0.5]{m/s}.
The depth range of the ZED camera is set to \unit[{$[0.3, 20]$}]{m}, and its FOV is $90^\circ$.
The measurement range of Hokuyo lidar is set to \unit[{$[0.1, 30]$}]{m}, its FOV is $270^\circ$, and its angular resolution is $0.25^\circ$.

\subsubsection{Pedestrian Configuration}
\label{subsec:PEDSIM_configuration}
PEDSIM is a microscopic pedestrian crowd simulation library, which uses the social forces model \cite{helbing1995social, moussaid2009experimental} to guide the motion of individual pedestrians:
\begin{equation}
    \mathbf{F}_p = \mathbf{F}^{\rm des}_p + \mathbf{F}^{\rm obs}_p + \mathbf{F}^{\rm per}_p + \mathbf{F}^{\rm rob}_p,
    \label{eq:sf_model}
\end{equation}
where $\mathbf{F}_p$ is the resultant force that determines the motion of a pedestrian $p$;
$\mathbf{F}^{\rm des}_p$ pulls a pedestrian towards a destination; 
$\mathbf{F}^{\rm obs}_p$ pushes a pedestrian away from static obstacles; 
$\mathbf{F}^{\rm per}_p$ models interactions with other pedestrians (\eg collision avoidance or grouping); and
$\mathbf{F}^{\rm rob}_p$ pushes pedestrians away from the robot, modeling the way people would naturally avoid collisions and thereby allowing our control policy to learn this behavior.
The first three terms are part of the standard model while the last term is new.

\subsubsection{Simulation Configuration}
\label{subsec:simulation_configuration}
Figure~\ref{fig:lobby} shows the main Gazebo simulation environment, a replica of the lobby in the College of Engineering building at Temple University, that we used to train and test our control policies.
This environment is roughly \unit[$25\times10$]{m} in size and is filled with a number of static obstacles (\eg chairs, tables, a security desk, waste bins, and pillars).
The second environment, shown in Fig.~\ref{fig:autolab}, is a replica of the Autolab building that is about \unit[$20\times10$]{m} in size.
The third environment, shown in Fig.~\ref{fig:cumberland}, is a replica of the Cumberland building that is about \unit[$20\times10$]{m} in size.
The fourth environment, shown in Fig.~\ref{fig:cumberland}, is a replica of the Freiburg building that is about \unit[$20\times10$]{m} in size.
The fifth environment, shown in Fig.~\ref{fig:square}, is an artificially created environment that is about \unit[$20\times20$]{m} in size.
Using these five worlds, we set up six types of scenarios:
\begin{enumerate}
    \item \textbf{Lobby world with 34 pedestrians}: train our DRL control policies
    \item \textbf{Lobby world with 5-55 pedestrians}: test generalization across different crowd densities
    \item \textbf{Autolab world with 25-35 pedestrians}: test generalization to unseen environments/crowd densities
    \item \textbf{Cumberland world with 25-35 pedestrians}: test generalization to unseen environments/crowd densities
    \item \textbf{Square world with 25-35 pedestrians}: test generalization to unseen environments/crowd densities
    \item \textbf{Freiburg world with 25-35 pedestrians}: test generalization to unseen environments/crowd densities
\end{enumerate}

We use an Nvidia DGX-1 server with 40 CPU cores, 8x Tesla V100 GPUs with \unit[16]{GB} memory, and \unit[512]{GB} of RAM to train our DRL networks. 
We use a desktop computer with an Intel i7-6800K CPU, a GeForce GTX 1080 GPU with \unit[8]{GB} memory, and \unit[16]{GB} of RAM to run all the simulations.
Both the training server and desktop computer run Ubuntu 20.04, ROS Noetic Ninjemys, and Gazebo 11.5.1.

\subsubsection{Hardware Configuration}
\label{subsec:hardware_configuration}
In the hardware tests, the Turtlebot2 is equipped with an NVIDIA Jetson AVG Xavier embedded computer containing a cluster of 8 Carmel ARM cores, a 512 Core Volta GPU, and \unit[16]{GB} RAM memory.
The Xavier runs Ubuntu 18.04 system and ROS Melodic Morenia, and we use the \unit[30]{W} power mode.

\begin{figure}[t]
    \centering
    \subfigure[Indoor hallway]{
            \centering
            \includegraphics[width=0.9\linewidth]{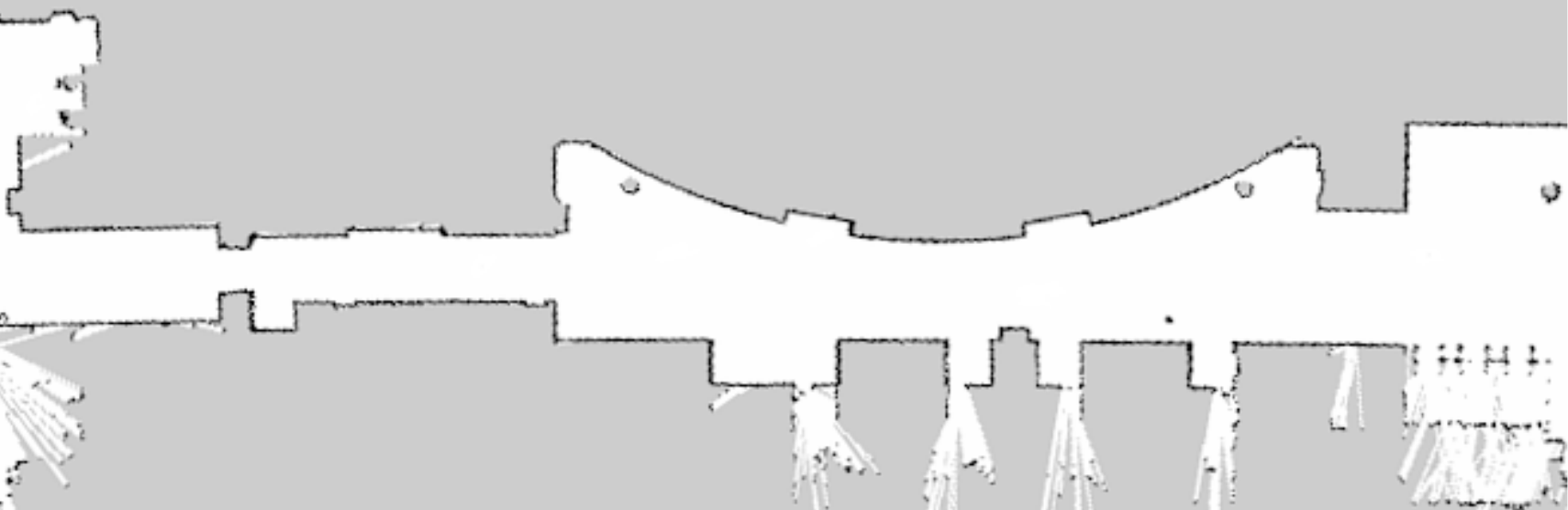}
            \label{fig:indoor_hallway}
    }%
    \vspace{0.01cm}
    \subfigure[Indoor lobby]{
            \centering
            \includegraphics[width=0.344\linewidth]{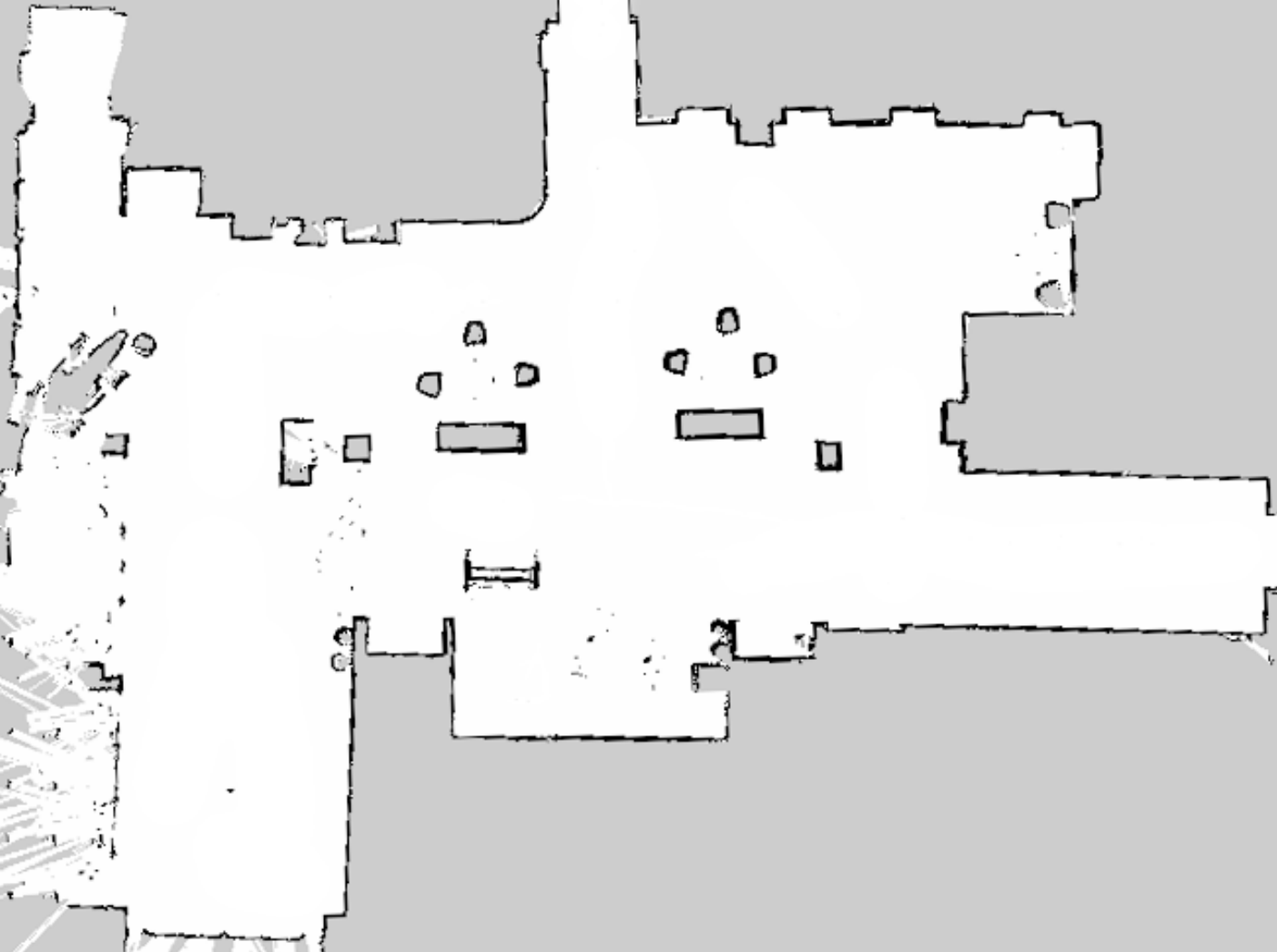}
            \label{fig:indoor_lobby}
    }%
    \subfigure[Outdoor hallway]{
            \centering
            \includegraphics[width=0.61\linewidth]{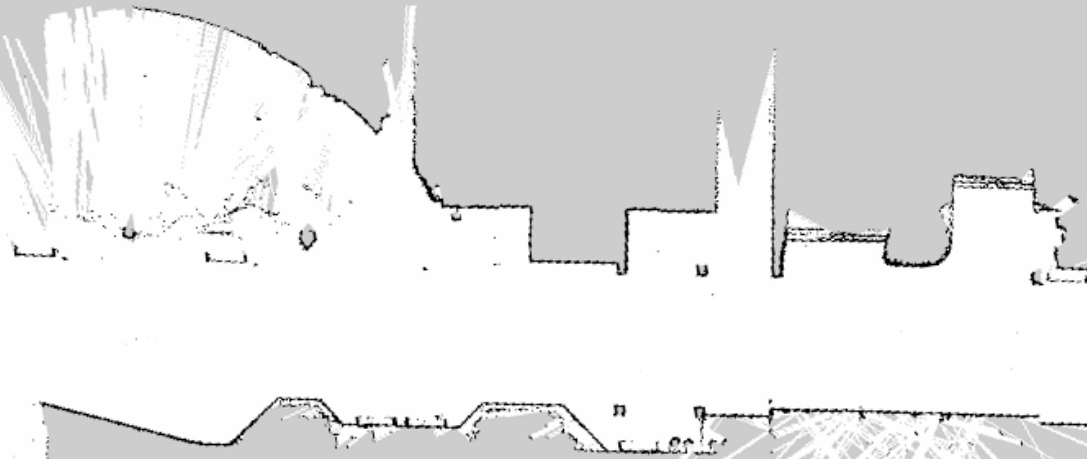}
            \label{fig:outdoor_hallway}
    }%
    \caption{Floorplans of real-world environments. 
             All of these three environments are at Temple University. 
    }
    \label{fig:floorplan}
\end{figure}

We test our robot in three environments, shown in Fig.~\ref{fig:floorplan}.
Figure~\ref{fig:indoor_hallway} is the indoor hallway connecting the Engineering building and Science building, which is about \unit[$66\times5$]{m} in size.
Figure~\ref{fig:indoor_lobby} is the indoor structured lobby in the Engineering building, which is about \unit[$20\times10$]{m} in size.
This is the same environment that we replicated in Gazebo to use for training our policy.
Figure~\ref{fig:outdoor_hallway} is the outdoor hallway outside of the Science building, which is about \unit[$60\times8$]{m} in size.

We conduct two sets of hardware experiments.
First, we test the real-world applicability in different crowd sizes using the indoor hallway environment (Fig.~\ref{fig:indoor_hallway}).
Second, we test the applicability in different environments using all three test environments.

\subsection{Training Procedure}
\label{sec:training_procedure}
We use the stable-baselines3 framework \cite{stable-baselines3} to implement the PPO algorithm and train our DRL networks.
We use the step decay with an initial learning rate of $10^{-3}$ as the learning rate schedule and set the mini-batch size to 512 during the training process.
To train our control policies, we use the Lobby environment (Fig.~\ref{fig:lobby}) with 34 pedestrians in it.
The robot is then repeatedly assigned to reach a random goal from a random start position within the free space of the map.
We used this procedure to train two different DRL-based control policies, one using the full reward \eqref{eq:reward} (DRL-VO) and one that does not use the heading direction reward $r_d^t$ (DRL).

\subsection{Simulation Results}
\label{subsec:simulation_results}
\begin{table*}[t]
    \small\sf\centering
    \caption{Navigation results at different crowd densities}
    \scalebox{0.9}{
    \begin{tabular}{l l l l l l}
        \toprule
        \textbf{Environment} & \textbf{Method} & \textbf{Success Rate} & \textbf{Average Time (s)} & \textbf{Average Length (m)} & \textbf{Average Speed (m/s)} \\ 
        \midrule
        \multirow{6}{*}{\begin{tabular}[c]{@{}c@{}}Lobby world,\\ 5 pedestrians\end{tabular}} 
         & DWA \cite{fox1997dynamic} & 0.93 & 12.10 & \textbf{5.08} & 0.42  \\  
         & CNN \cite{xie2021towards} & - & - & - & -  \\  
         & A1-RD \cite{guldenring2020learning} & - & - & - & -  \\  
         & A1-RC \cite{guldenring2020learning} & 0.91 & 14.11 & 6.14 & 0.44  \\  
         & DRL & 0.84 & 13.22 & 6.08 & 0.46  \\  
         & DRL-VO & \textbf{0.94} & \textbf{11.41} & 5.28 & \textbf{0.46}  \\ \hline
        \multirow{6}{*}{\begin{tabular}[c]{@{}c@{}}Lobby world,\\ 15 pedestrians\end{tabular}} 
         & DWA \cite{fox1997dynamic} & 0.94 & 12.21 & \textbf{5.09} & 0.42  \\  
         & CNN \cite{xie2021towards} & - & - & - & -  \\  
         & A1-RD \cite{guldenring2020learning} & 0.94 & 15.69 & 5.78 & 0.37  \\  
         & A1-RC \cite{guldenring2020learning} & 0.94 & 14.36 & 6.40 & 0.45  \\  
         & DRL & 0.80 & 13.66 & 6.24 & 0.46  \\  
         & DRL-VO  & \textbf{0.95} & \textbf{11.34} & 5.26 & \textbf{0.46} \\ \hline
        \multirow{6}{*}{\begin{tabular}[c]{@{}c@{}}Lobby world,\\ 25 pedestrians\end{tabular}} 
         & DWA \cite{fox1997dynamic} & 0.82 & 13.49 & \textbf{5.12} & 0.38  \\  
         & CNN \cite{xie2021towards} & 0.80 & 19.31 & 6.16 & 0.32  \\  
         & A1-RD \cite{guldenring2020learning} & 0.90 & 17.87 & 6.07 & 0.34  \\  
         & A1-RC \cite{guldenring2020learning} & 0.88 & 14.18 & 6.26 & 0.44  \\  
         & DRL & 0.81 & 13.73 & 6.24 & 0.45  \\  
         & DRL-VO & \textbf{0.92} & \textbf{11.37} & 5.29 & \textbf{0.47}  \\ \hline
        \multirow{6}{*}{\begin{tabular}[c]{@{}c@{}}Lobby world, \\ 35 pedestrians\end{tabular}} 
         & DWA \cite{fox1997dynamic} & 0.82 & 14.18 & \textbf{5.15} & 0.36  \\  
         & CNN \cite{xie2021towards} & 0.81 & 14.30 & 5.40 & 0.38  \\  
         & A1-RD \cite{guldenring2020learning} & 0.86 & 17.82 & 6.06 & 0.34  \\  
         & A1-RC \cite{guldenring2020learning} & 0.77 & 16.81 & 6.89 & 0.41  \\  
         & DRL & 0.75 & 14.30 & 6.46 & 0.45  \\  
         & DRL-VO  & \textbf{0.88} & \textbf{11.42} & 5.31 & \textbf{0.46}  \\ \hline
         \multirow{6}{*}{\begin{tabular}[c]{@{}c@{}}Lobby world, \\ 45 pedestrians\end{tabular}} 
         & DWA \cite{fox1997dynamic} & 0.77 & 15.39 & \textbf{5.16} & 0.34  \\  
         & CNN \cite{xie2021towards} & 0.79 & 16.65 & 5.62 & 0.34  \\  
         & A1-RD \cite{guldenring2020learning} & 0.76 & 23.16 & 6.61 & 0.29  \\  
         & A1-RC \cite{guldenring2020learning} & 0.77 & 14.65 & 6.28 & 0.43  \\  
         & DRL & 0.69 & 13.96 & 6.41 & 0.46  \\  
         & DRL-VO  & \textbf{0.81} & \textbf{11.65} & 5.37 & \textbf{0.46}  \\ \hline
        \multirow{6}{*}{\begin{tabular}[c]{@{}c@{}}Lobby world, \\ 55 pedestrians\end{tabular}} 
         & DWA \cite{fox1997dynamic} & 0.67 & 16.05 & \textbf{5.22} & 0.33   \\  
         & CNN \cite{xie2021towards} & 0.70 & 19.13 & 5.94 & 0.31  \\  
         & A1-RD \cite{guldenring2020learning} & 0.67 & 26.90 & 6.58 & 0.25  \\  
         & A1-RC \cite{guldenring2020learning} & 0.66 & 16.36 & 6.73 & 0.41  \\  
         & DRL & 0.60 & 13.39 & 6.13 & 0.46  \\  
         & DRL-VO  & \textbf{0.79} & \textbf{11.76} & 5.39 & \textbf{0.46}  \\ 
         \bottomrule
    \end{tabular}
    }
    \label{tab:densities}
\end{table*}

\begin{table*}[t]
    \small\sf\centering
    \caption{Navigation results at different unseen environments with 25 pedestrians}
    \scalebox{0.9}{
    \begin{tabular}{l l l l l l}
        \toprule
        \textbf{Environment} & \textbf{Method} & \textbf{Success Rate} & \textbf{Average Time (s)} & \textbf{Average Length (m)} & \textbf{Average Speed (m/s)} \\ 
        \midrule
        \multirow{6}{*}{\begin{tabular}[c]{@{}c@{}}Lobby world,\\ 25 pedestrians\end{tabular}} 
         & DWA \cite{fox1997dynamic} & 0.82 & 13.49 & \textbf{5.12} & 0.38  \\ 
         & CNN \cite{xie2021towards} & 0.80 & 19.31 & 6.16 & 0.32  \\ 
         & A1-RD \cite{guldenring2020learning} & 0.90 & 17.87 & 6.07 & 0.34  \\ 
         & A1-RC \cite{guldenring2020learning} & 0.88 & 14.18 & 6.26 & 0.44  \\ 
         & DRL & 0.81 & 13.73 & 6.24 & 0.45  \\ 
         & DRL-VO & \textbf{0.92} & \textbf{11.37} & 5.29 & \textbf{0.47}  \\ \hline
        \multirow{6}{*}{\begin{tabular}[c]{@{}c@{}}Autolab world,\\  25 pedestrians\end{tabular}} 
         & DWA \cite{fox1997dynamic} & 0.88 & 13.50 & \textbf{5.47} & 0.41  \\ 
         & CNN \cite{xie2021towards} & 0.78 & 21.59 & 6.61 & 0.31  \\ 
         & A1-RD \cite{guldenring2020learning} & 0.88 & 18.92 & 6.54 & 0.35  \\ 
         & A1-RC \cite{guldenring2020learning} & 0.77 & 15.63 & 7.35 & 0.43  \\ 
         & DRL & 0.75 & 15.66 & 7.03 & 0.45  \\ 
         & DRL-VO  & \textbf{0.91} & \textbf{12.79} & 5.97 & \textbf{0.47} \\ \hline
        \multirow{6}{*}{\begin{tabular}[c]{@{}c@{}}Cumberland world,\\ 25 pedestrians\end{tabular}} 
         & DWA \cite{fox1997dynamic} & 0.78 & 15.16 & \textbf{5.53} & 0.37  \\ 
         & CNN \cite{xie2021towards} & 0.63 & 27.80 & 7.73 & 0.28  \\ 
         & A1-RD \cite{guldenring2020learning} & \textbf{0.91} & 17.47 & 6.63 & 0.38  \\ 
         & A1-RC \cite{guldenring2020learning} & 0.81 & 16.31 & 6.93 & 0.43  \\ 
         & DRL & 0.66 & 14.28 & 6.38 & 0.45  \\  
         & DRL-VO & 0.87 & \textbf{12.42} & 5.78 & \textbf{0.47}  \\ 
         \hline
        \multirow{6}{*}{\begin{tabular}[c]{@{}c@{}}Freiburg world, \\ 25 pedestrians\end{tabular}} 
         & DWA \cite{fox1997dynamic} & 0.83 & 13.90 & \textbf{5.75} & 0.41  \\  
         & CNN \cite{xie2021towards} & 0.48 & 30.38 & 8.30 & 0.27  \\  
         & A1-RD \cite{guldenring2020learning} & 0.81 & 17.95 & 6.61 & 0.37  \\  
         & A1-RC \cite{guldenring2020learning} & 0.74 & 17.06 & 7.35 & 0.43  \\  
         & DRL & 0.53 & 15.68 & 7.06 & 0.45  \\  
         & DRL-VO  & \textbf{0.84} & \textbf{12.98} & 5.93 & \textbf{0.46}  \\ \hline
        \multirow{6}{*}{\begin{tabular}[c]{@{}c@{}}Square world, \\ 25 pedestrians\end{tabular}} 
         & DWA \cite{fox1997dynamic} & 0.94 & 19.62 & \textbf{8.48} & 0.43   \\  
         & CNN \cite{xie2021towards} & 0.65 & 53.35 & 14.72 & 0.28  \\  
         & A1-RD \cite{guldenring2020learning} & \textbf{0.95} & 21.39 & 9.05 & 0.42  \\  
         & A1-RC \cite{guldenring2020learning} & 0.89 & 21.44 & 9.74 & 0.45  \\  
         & DRL & 0.83 & 22.62 & 10.77 & \textbf{0.48}  \\  
         & DRL-VO  & 0.86 & \textbf{18.19} & 8.61 & 0.47  \\ 
         \bottomrule
    \end{tabular}
    }
    \label{tab:environments25}
\end{table*}

\begin{table*}[t]
    \small\sf\centering
    \caption{Navigation results at different unseen environments with 35 pedestrians}
    \scalebox{0.9}{
    \begin{tabular}{l l l l l l}
        \toprule
        \textbf{Environment} & \textbf{Method} & \textbf{Success Rate} & \textbf{Average Time (s)} & \textbf{Average Length (m)} & \textbf{Average Speed (m/s)} \\ 
        \midrule
        \multirow{6}{*}{\begin{tabular}[c]{@{}c@{}}Lobby world,\\ 35 pedestrians\end{tabular}} 
         & DWA \cite{fox1997dynamic} & 0.82 & 14.18 & \textbf{5.15} & 0.36  \\ 
         & CNN \cite{xie2021towards} & 0.81 & 14.30 & 5.40 & 0.38  \\ 
         & A1-RD \cite{guldenring2020learning} & 0.86 & 17.82 & 6.06 & 0.34  \\ 
         & A1-RC \cite{guldenring2020learning} & 0.77 & 16.81 & 6.89 & 0.41  \\ 
         & DRL & 0.75 & 14.30 & 6.46 & 0.45  \\ \
         & DRL-VO  & \textbf{0.88} & \textbf{11.42} & 5.31 & \textbf{0.46}  \\ 
         \hline
        \multirow{6}{*}{\begin{tabular}[c]{@{}c@{}}Autolab world,\\ 35 pedestrians\end{tabular}} 
         & DWA \cite{fox1997dynamic} & 0.81 & 16.52 & \textbf{5.55} & 0.34  \\ 
         & CNN \cite{xie2021towards} & 0.68 & 22.73 & 6.65 & 0.29  \\
         & A1-RD \cite{guldenring2020learning} & - & - & - & -  \\ 
         & A1-RC \cite{guldenring2020learning} & 0.73 & 17.11 & 6.93 & 0.41  \\ 
         & DRL & 0.59 & 14.40 & 7.01 & 0.46  \\ 
         & DRL-VO  & \textbf{0.84} & \textbf{12.88} & 5.98 & \textbf{0.46} \\ 
         \hline
        \multirow{6}{*}{\begin{tabular}[c]{@{}c@{}}Cumberland world,\\ 35 pedestrians\end{tabular}} 
         & DWA \cite{fox1997dynamic} & 0.74 & 16.28 & \textbf{5.57} & 0.34  \\ 
         & CNN \cite{xie2021towards} & 0.60 & 24.25 & 7.08 & 0.29  \\ 
         & A1-RD \cite{guldenring2020learning} & \textbf{0.88} & 18.04 & 6.69 & 0.37 \\
         & A1-RC \cite{guldenring2020learning} & 0.77 & 15.37 & 6.66 & 0.43  \\ 
         & DRL & 0.56 & 15.79 & 6.97 & 0.44  \\ 
         & DRL-VO & 0.78 & \textbf{12.62} & 5.84 & \textbf{0.46}  \\ 
         \hline
        \multirow{6}{*}{\begin{tabular}[c]{@{}c@{}}Freiburg world, \\ 35 pedestrians\end{tabular}} 
         & DWA \cite{fox1997dynamic} & 0.70 & 18.15 & \textbf{5.78} & 0.32  \\ 
         & CNN \cite{xie2021towards} & 0.57 & 20.09 & 6.66 & 0.33  \\ 
         & A1-RD \cite{guldenring2020learning} & - & - & - & -  \\ 
         & A1-RC \cite{guldenring2020learning} & 0.65 & 17.11 & 6.93 & 0.41  \\ 
         & DRL & 0.39 & 18.21 & 7.61 & 0.42  \\ 
         & DRL-VO  & \textbf{0.76} & \textbf{13.37} & 6.16 & \textbf{0.46}  \\ \hline
        \multirow{6}{*}{\begin{tabular}[c]{@{}c@{}}Square world, \\ 35 pedestrians\end{tabular}} 
         & DWA \cite{fox1997dynamic} & 0.93 & 19.81 & \textbf{8.51} & 0.43   \\ 
         & CNN \cite{xie2021towards} & 0.85 & 23.81 & 9.23 & 0.39  \\ 
         & A1-RD \cite{guldenring2020learning} & \textbf{0.96} & 21.85 & 9.06 & 0.41  \\ 
         & A1-RC \cite{guldenring2020learning} & 0.89 & 21.29 & 9.67 & 0.45  \\ 
         & DRL & 0.86 & 21.48 & 10.27 & \textbf{0.48}  \\ 
         & DRL-VO  & 0.87 & \textbf{18.32} & 8.66 & 0.47  \\ 
         \bottomrule
    \end{tabular}
    }
    \label{tab:environments35}
\end{table*}

We test these two DRL-based policies (\ie DRL and DRL-VO), along with other's DRL-based policies \cite{guldenring2020learning} (\ie A1-RD and A1-RC), the CNN-based policy \cite{xie2021towards}, and the DWA planner \cite{fox1997dynamic}, in each of the last five types of scenarios from Sec.~\ref{subsec:simulation_configuration}.
Note that we use these baseline policies directly from their papers \cite{guldenring2020learning, xie2021towards, fox1997dynamic} without any retraining or parameter tuning. 
We compare our algorithms to these baselines using four commonly used metrics from the navigation literature: 
\begin{enumerate}
    \item \textbf{Success rate}: the fraction of collision-free trials
    
    \item \textbf{Average time}: the average travel time
    
    \item \textbf{Average length}: the average distance traveled
    
    \item \textbf{Average speed}: the average travel speed
\end{enumerate}

We run four trials for each experimental configuration and control policy, where each trial consists of the robot navigating through a predefined series of 25 goal points around the corresponding test environment.
Note that we use the \texttt{amcl} ROS package to provide the robot localization service in known maps. 
All of these four trials use the same initial conditions, however, the results vary due to randomized sensor noise and variations in the pedestrian motion from the social force model.

\subsubsection{Different Crowd Densities}
\label{subsubsec:different_crowd_densities}
We first focus on the Lobby environment with 5-55 (sampling interval 10) pedestrians used for testing the generalization across different crowd sizes.
Table~\ref{tab:densities} presents the results obtained from these trials, where we observe three key phenomena.
First, our DRL-VO policy has a much higher success rate than our DRL policy in each crowd size, while having a similar average speed.
This shows that the proposed VO-based heading direction reward is beneficial and plays a key role in enabling the robot to maintain a good balance between collision avoidance and speed.
Second, compared with other control policies, our DRL-VO policy has the highest success rate and the fastest average speed in every situation, with these differences growing as the density of pedestrians increases.
This indicates that our policy is able to better generalize than the other approaches, especially compared to the CNN-based policy \cite{xie2021towards}, which could not even reach the goal in some situations.
Third, the average speed of the robot using our DRL-VO policy remains nearly constant across all situations, regardless of crowd density, allowing it to reach the goal most quickly.
Compared to DWA \cite{fox1997dynamic}, we see that our planner takes a slightly longer route to the goal, indicating a preference for giving a wider berth to pedestrians.

\subsubsection{Different Unseen Environments}
\label{subsubsec:different_unseen_environments_with_25_pedestrians}
We next consider the navigation behavior of our DRL-VO policy in different unseen environments, namely the Autolab, Cumberland, Freiburg, and Square environments from Fig.~\ref{fig:gazebo}.
In each environment, we test two different crowd densities: 25 and 35 pedestrians.
Tables~\ref{tab:environments25} and~\ref{tab:environments35} summarize these results.
As can be seen, our novel DRL-VO policy still has both a higher success rate and slightly higher average speed than our DRL policy, with the only exception being a slightly lower speed in Square world. 
This further demonstrates the utility of our VO-based reward term to learn a more robust navigation policy and achieve a better trade-off between collision avoidance and speed.

Our DRL-VO policy also has the highest success rate out of all the algorithms in the lobby, Autolab, and Freiburg worlds.
Additionally, our policy yields a faster and more consistent speed than other policies and tends to have the shortest path, aside from the DWA planner.
This indicates that our policy has strong generalizability to different unseen environments.

The success rate of our policy drops, however, both in absolute and relative terms, in the Cumberland and Square worlds.
Both of these environments have larger open spaces compared to the rest in Fig.~\ref{fig:gazebo}.
Given this, two potential causes of this decrease in success rate are: 1) the policy was trained in the more structured lobby and so adding in more varied environments during training would further improve the generalizability or 2) our DRL-VO policy with VO-based reward function is more suitable for structured environments rather than open spaces.

The A1-RD policy \cite{guldenring2020learning} has the highest success rate in the Cumberland and Square environments.
We found this to be primarily due to its ``stop and wait'' strategy, where a robot that detects an obstacle in front of it will turn to a free space, stop, and wait for the obstacles to disappear.
However, this behavior is the same regardless of the number of obstacles or whether these obstacles are static or dynamic.
This can cause the robot to become frozen when it encounters a ``C-shaped'' obstacle because its policy only used raw lidar data and it was trained using a typical passive collision avoidance strategy based only on distance.
This causes a robot using A1-RD to fail to reach the goal in both the Autolab and Freiburg environments with 35 pedestrians.
In contrast to this, our DRL-VO policy can distinguish between static obstacles and dynamic pedestrians using the combination of lidar data and pedestrian kinematics.
Moreover, depending on the crowd densities and velocity obstacle space, our DRL-VO policy will always tune its heading direction to actively avoid pedestrians and move towards the goal point. 
See the videos in Multimedia Extension 1 to see these differences in the behavior of A1-RD and DRL-VO detailed.

\begin{figure*}[t]
    \centering
    \subfigure[Linear velocity distribution]
    {
        \centering
        \includegraphics[width=0.45\textwidth]{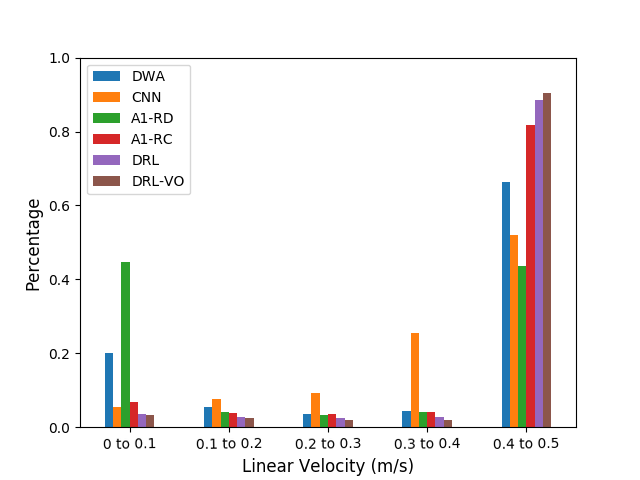}
        \label{fig:linear_velocity_distribution}
    } 
    \subfigure[Angular velocity distribution]
    {
        \centering
        \includegraphics[width=0.45\textwidth]{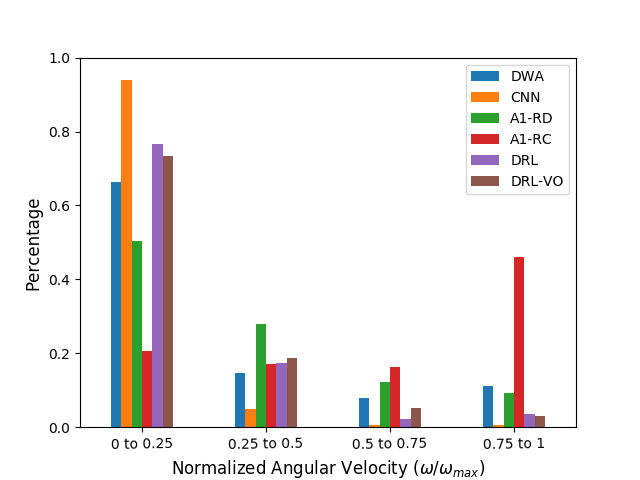}
        \label{fig:angular_velocity_distribution}
    }%
   
    \caption{
            Histograms of the output velocities for different control policies. 
            Data was collected from the lobby simulation world with 35 pedestrians.
    }
    \label{fig:velocity_distribution}
\end{figure*}

\subsubsection{Output Velocity Distribution}
\label{subsubsec:output_velocity_distribution}
We also analyze the distribution of command velocities from each tested control policy to better understand why our DRL-VO policy has a faster average speed than other model-based and learning-based control policies. 
To do this we make histograms of the linear and angular velocity, using bins of size \unit[0.1]{m/s} and \unit[0.25]{$\omega_{\rm max}$}, respectively.
Note that we take the absolute value and normalize the angular velocity to ignore the direction of the turn and to account for different maximum values across the different control policies.

We observe two interesting trends in the data in Fig.~\ref{fig:velocity_distribution}.
First, the majority of the linear velocities generated by all policies lie at either extreme of the range.
Our DRL and DRL-VO policies have the highest percentage at the upper end of the range (about 90\%), which is why we achieve a consistently higher average speed.
On the other hand, the A1-RD policy has the highest percentage in the lower end of the range (over 40\%), showing the prevalence of the ``stop and wait'' strategy noted above.
The only other policy with a significant percentage of the output near 0 is DWA, which occurs due to deadlock in particularly crowded regions.

Second, most of the control policies have similar angular velocity distributions, with a majority of commands being small changes to update the heading angle of the robot (\ie between \unit[{$[0,0.25]$}]{$\omega_{\rm max}$}).
However, there are some key differences.
First, the two outliers are the CNN policy, which yields very few commands outside of this minimum range, and the A1-RC policy, which generates a majority of commands at the upper range.
Our DRL and DRL-VO policies generate fewer commands with high angular velocities, which is a result of the path smoothness term in our reward function, $r_w$.

\subsubsection{Failure Cases and Solutions}
\label{subsubsec:failure_cases_and_solutions}
Although our proposed DRL-VO policy is robust to different crowd densities and unseen environments, there are still some failure cases. 
One is that it can become difficult to find a collision-free direction in very high crowd densities. 
The other is the unexpected sudden changes in pedestrian movement, such as a person suddenly stepping out of a crowd.
Our analysis of these failures is caused by three reasons: 1) our simulated pedestrians occasionally bump into one another due to the imperfect social force model,
2) there are occasionally missed pedestrian tracks in the MHT due to occlusion and ambiguous data association, and 3) the safety guarantees of VOs do not extend to the resulting learned policy because we cannot explicitly define collision constraints and apply them to the black box network.
There are many possible ways to alleviate these issues in future work, including continuing to train the policies (learned in simulation) in real-world environments, improving the robustness of the MHT, adding probabilistic predictions of the future state of the environment, adding a control barrier function as a safety constraint like \cite{chen2017obstacle}, or simply add a safety module as a backup: when the obstacle is too close to the robot, the safety module will override the learned policy to stop and rotate the robot until it finds a collision-free space (much like the ``stop and wait'' in the A1-RD policy).

\subsection{Hardware Results}
\label{subsec:hardware_results}
\begin{figure*}[t]
    \centering
    \subfigure[t]{
            \centering
            \includegraphics[width=0.235\textwidth]{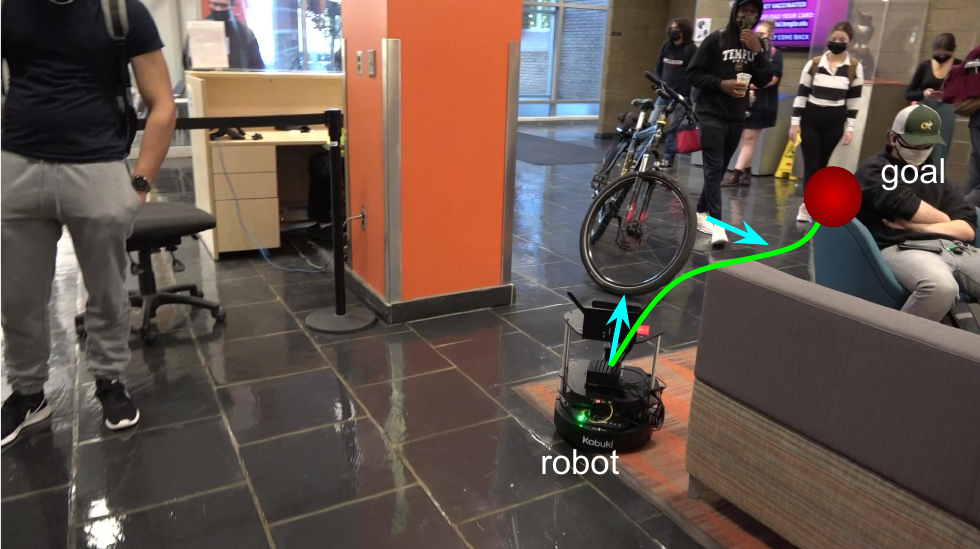}
            \label{fig:lobby_high_density1}
    }%
    \subfigure[t+1]{
            \centering
            \includegraphics[width=0.235\textwidth]{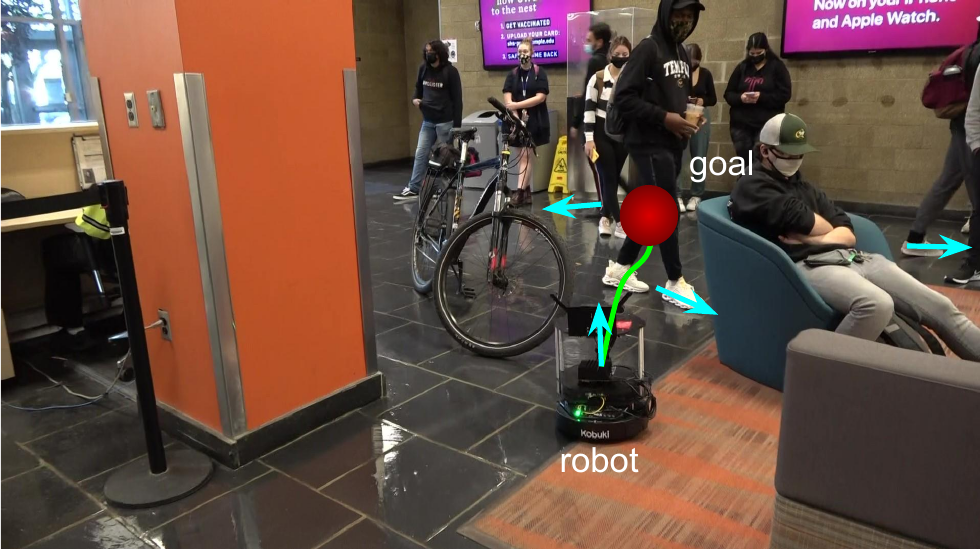}
            \label{fig:lobby_high_density2}
    }%
    \subfigure[t+2]{
            \centering
            \includegraphics[width=0.235\textwidth]{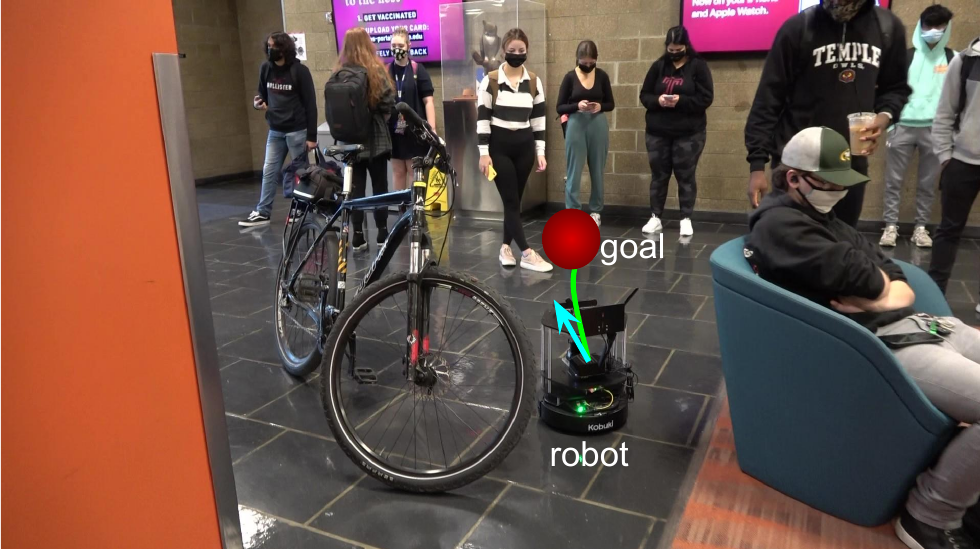}
            \label{fig:lobby_high_density3}
    }%
    \subfigure[t+3]{
            \centering
            \includegraphics[width=0.235\textwidth]{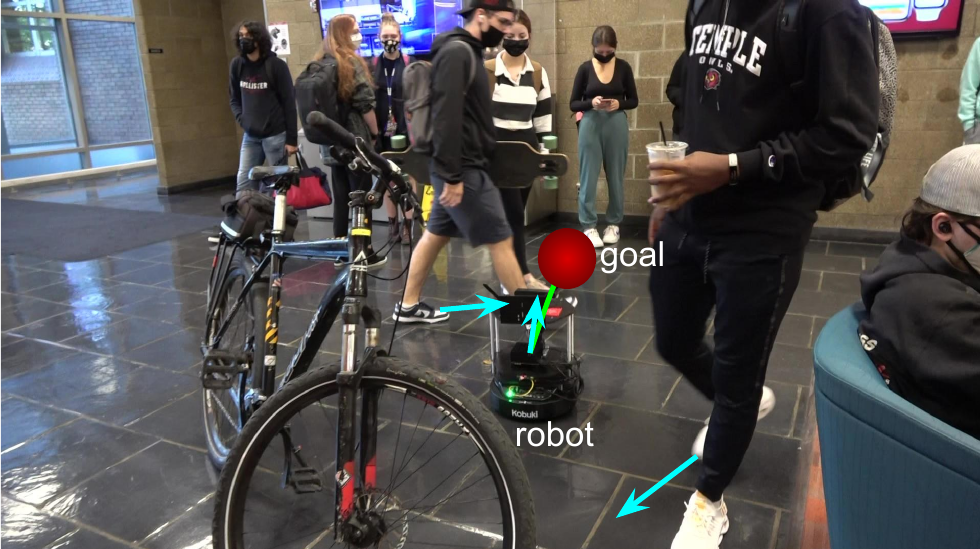}
            \label{fig:lobby_high_density4}
    }%
    \caption{Robot reactions to moving pedestrians in the indoor lobby with the high crowd density at different times.
            }
    \label{fig:lobby_high_density}
\end{figure*}

\begin{figure*}
    \centering
    \subfigure[t]{
            \centering
            \includegraphics[width=0.235\textwidth]{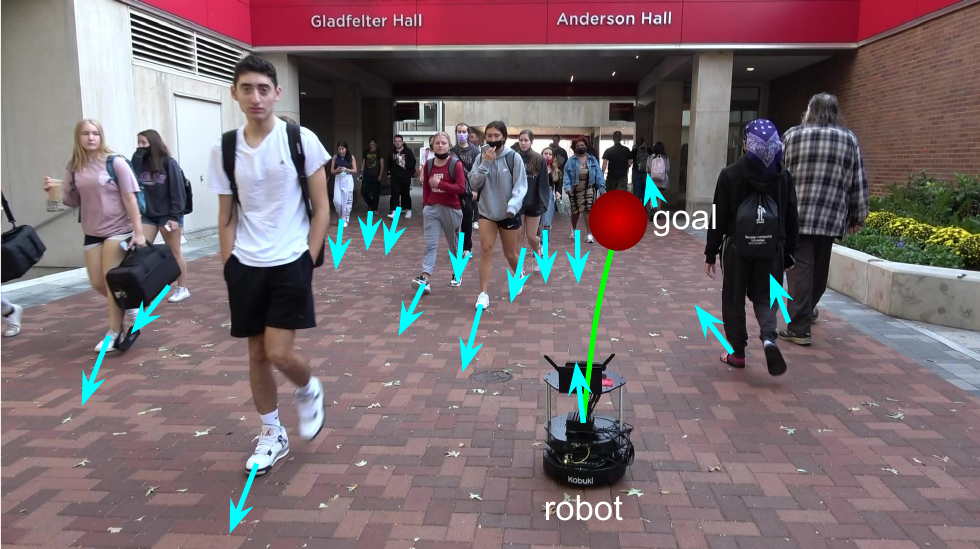}
            \label{fig:outdoor_hallway_high_density1}
    }%
    \subfigure[t+1]{
            \centering
            \includegraphics[width=0.235\textwidth]{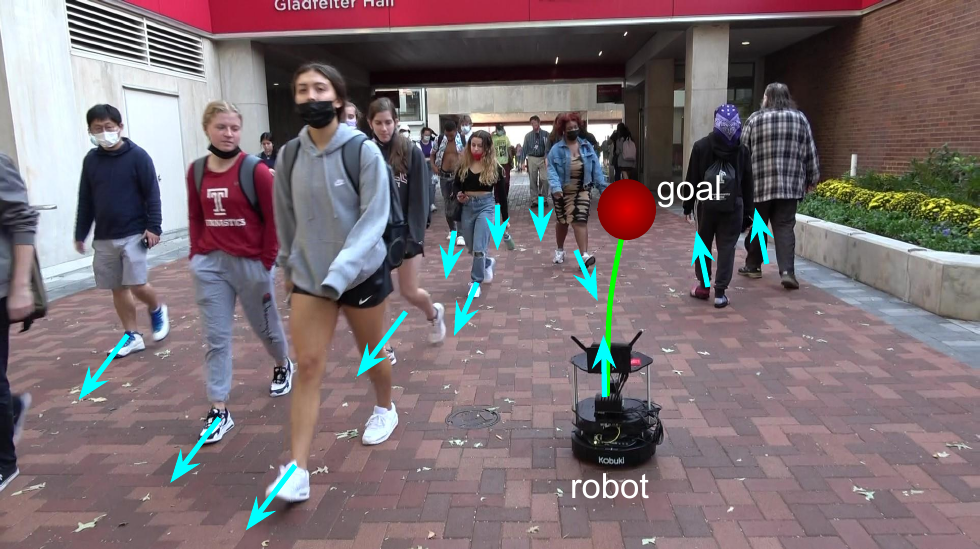}
            \label{fig:outdoor_hallway_high_density2}
    }%
    \subfigure[t+2]{
            \centering
            \includegraphics[width=0.235\textwidth]{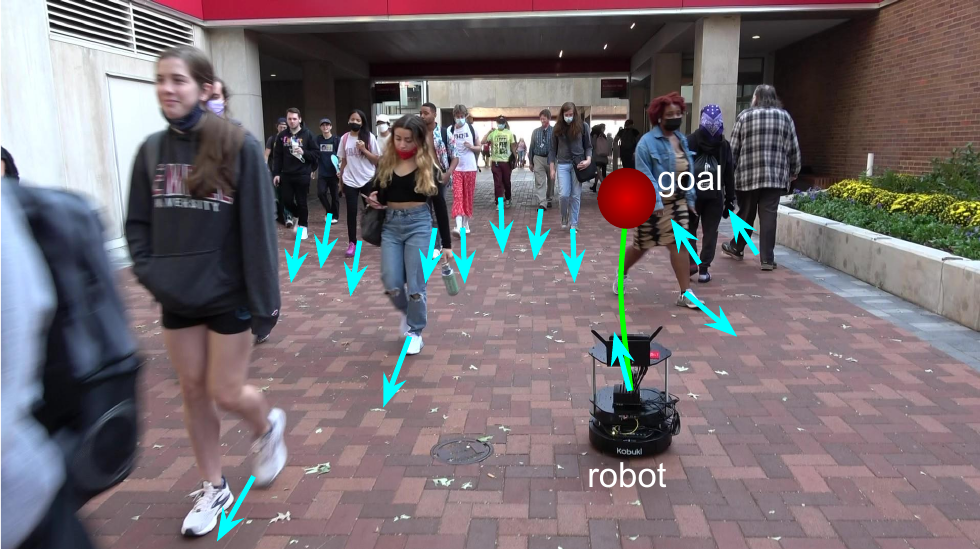}
            \label{fig:outdoor_hallway_high_density3}
    }%
    \subfigure[t+3]{
            \centering
            \includegraphics[width=0.235\textwidth]{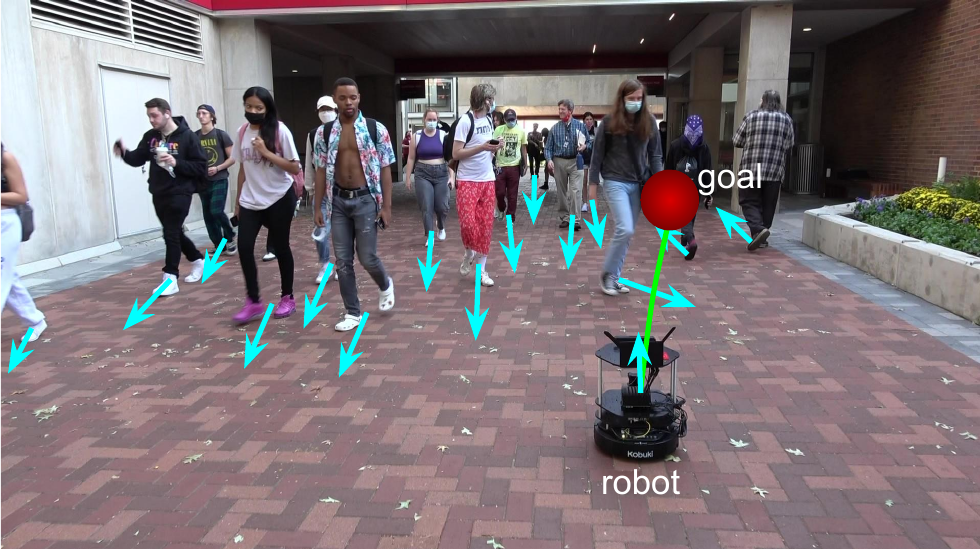}
            \label{fig:outdoor_hallway_high_density4}
    }%
    \caption{Robot reactions to moving pedestrians in the outdoor hallway with the high crowd density at different times.
            }
    \label{fig:outdoor_high_density}
\end{figure*}

Besides the simulated experiments, we also conduct a series of real-world experiments to demonstrate the applicability of our DRL-VO policy.
Instead of performing hardware experiments in the laboratory under carefully controlled conditions as \cite{guldenring2020learning, everett2021collision, chen2019crowd, liu2020decentralized, chen2020relational}, we focus on robotic navigation in real and difficult environments by testing our system during the peak periods of pedestrian traffic in the intervals between classes.
We test in three different parts of campus, an indoor hallway, an indoor lobby, and an outdoor hallway environment, shown in Fig.~\ref{fig:floorplan}.
In each environment, the robot passes through a predefined series of goal points with students of different crowd sizes naturally walking around and causing the robot to adjust its path, as Fig.~\ref{fig:hallway_low_density}-\ref{fig:outdoor_high_density} shows.

The real Turtlebot2 robot directly uses our DRL-VO policy trained from our Gazebo simulation platform without any fine-tuning.
The computational complexity of our DRL-VO policy is relatively low, with up to \unit[60]{FPS} inference speed on the Jetson AVG Xavier embedded computer, far above the 40 Hz of the sensors.
The space complexity of our DRL-VO policy is also low, with only \unit[32.137]{M} parameters.
This means our DRL-VO policy is clearly capable of real-time processing and is suitable for robots with limited resources.

\subsubsection{Different Crowd Densities}
\label{subsubsec:different_crowd_densities_hardware}
\begin{figure*}[t]
    \centering
    \subfigure[t]{
            \centering
            \includegraphics[width=0.235\textwidth]{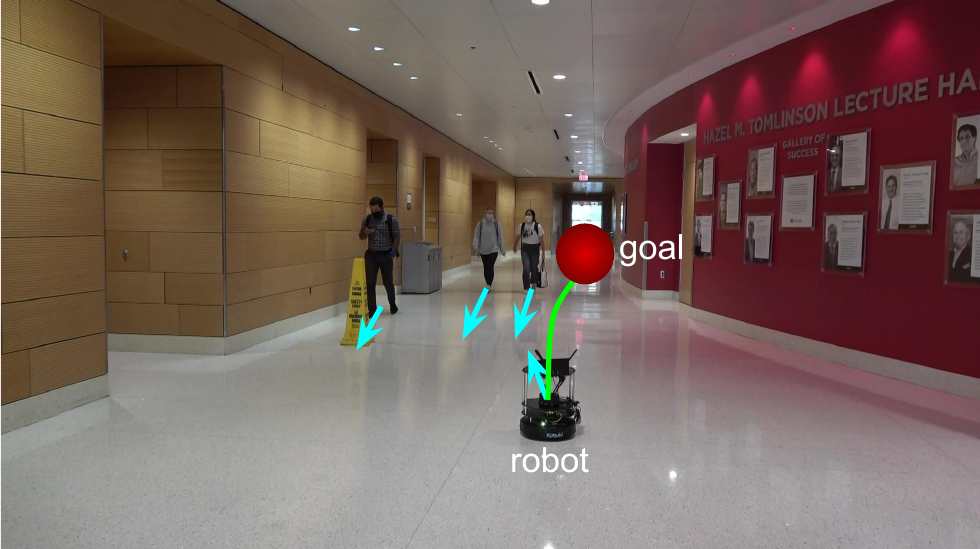}
            \label{fig:hallway_low_density1}
    }%
    \subfigure[t+1]{
            \centering
            \includegraphics[width=0.235\textwidth]{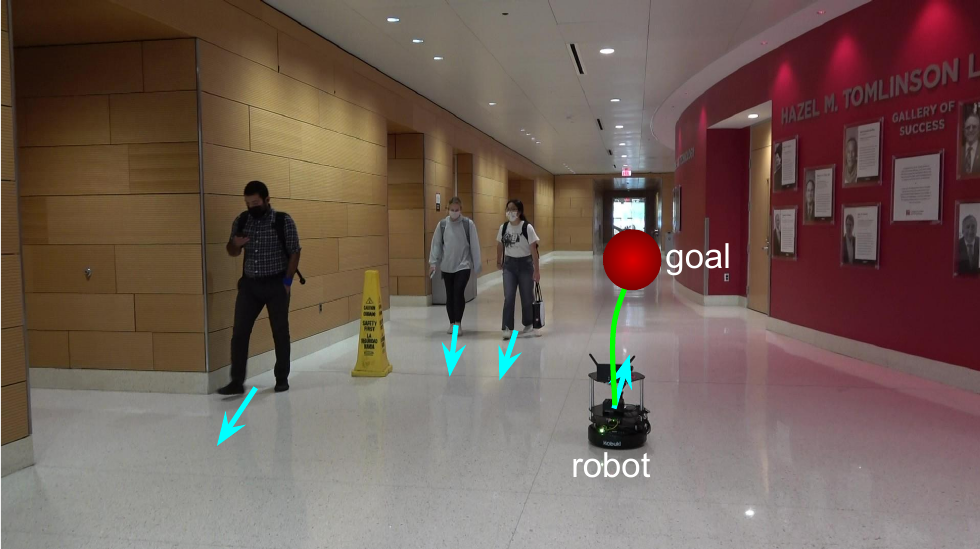}
            \label{fig:hallway_low_density2}
    }%
    \subfigure[t+2]{
            \centering
            \includegraphics[width=0.235\textwidth]{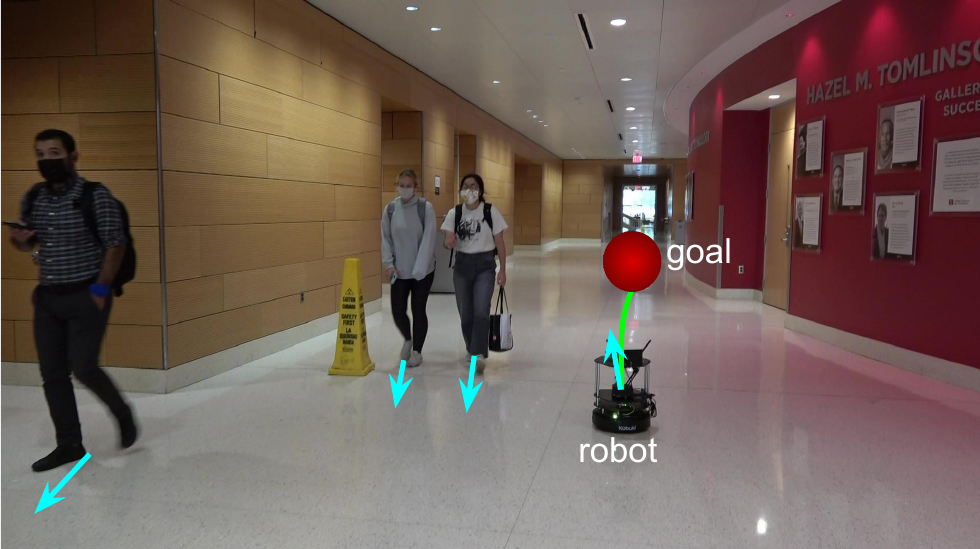}
            \label{fig:hallway_low_density3}
    }%
    \subfigure[t+3]{
            \centering
            \includegraphics[width=0.235\textwidth]{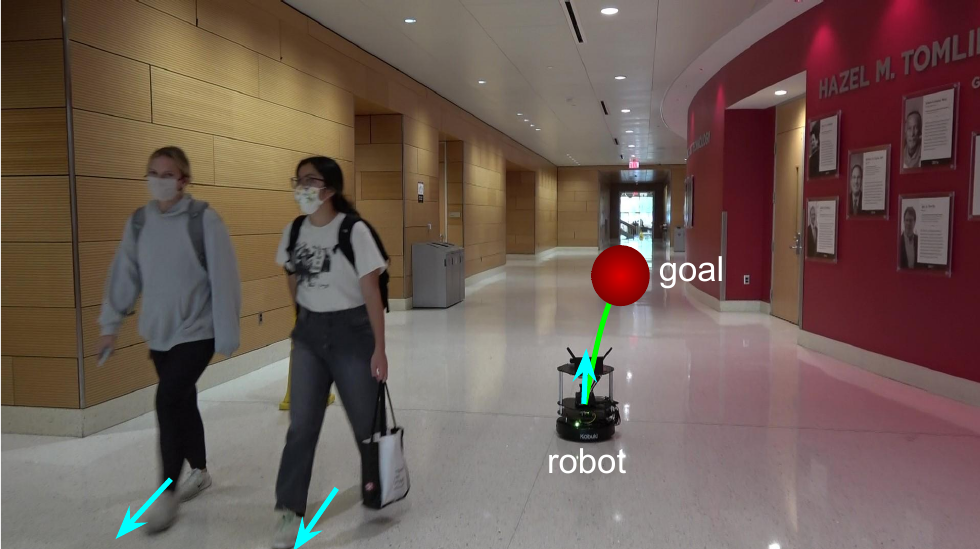}
            \label{fig:hallway_low_density4}
    }%
    \caption{Robot reactions to moving pedestrians in the indoor hallway with the low crowd density at different times.}
    \label{fig:hallway_low_density}
\end{figure*}

\begin{figure*}[t]
    \centering
    \subfigure[t]{
            \centering
            \includegraphics[width=0.235\textwidth]{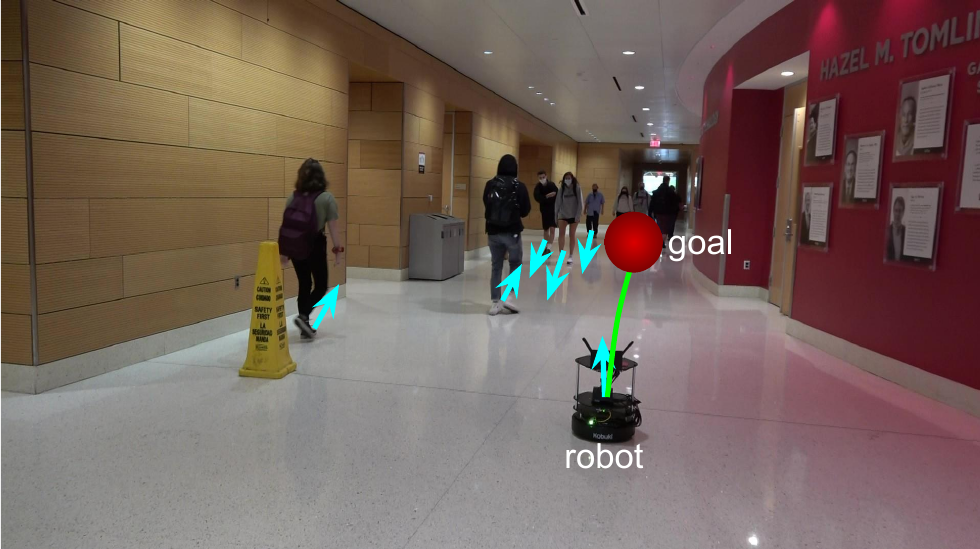}
            \label{fig:hallway_medium_density1}
    }%
    \subfigure[t+1]{
            \centering
            \includegraphics[width=0.235\textwidth]{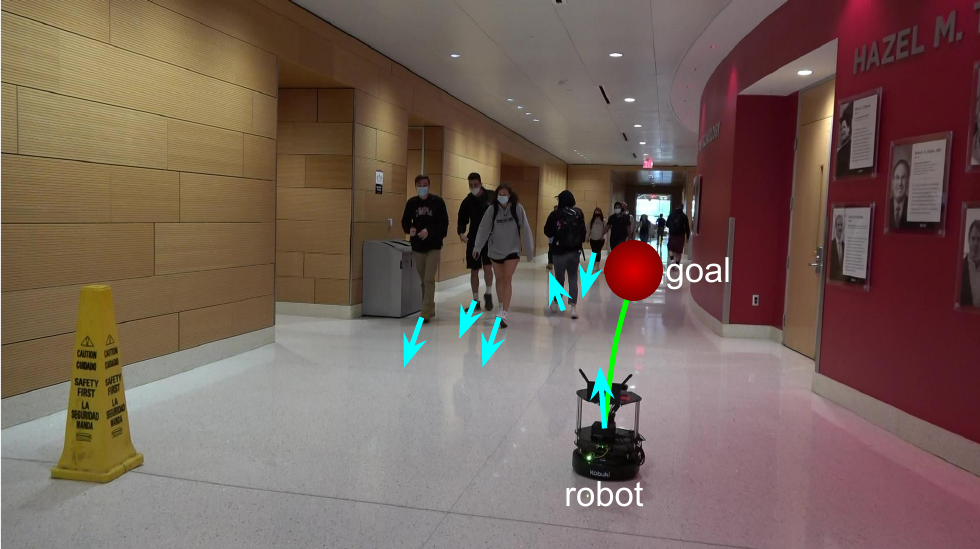}
            \label{fig:hallway_medium_density2}
    }%
    \subfigure[t+2]{
            \centering
            \includegraphics[width=0.235\textwidth]{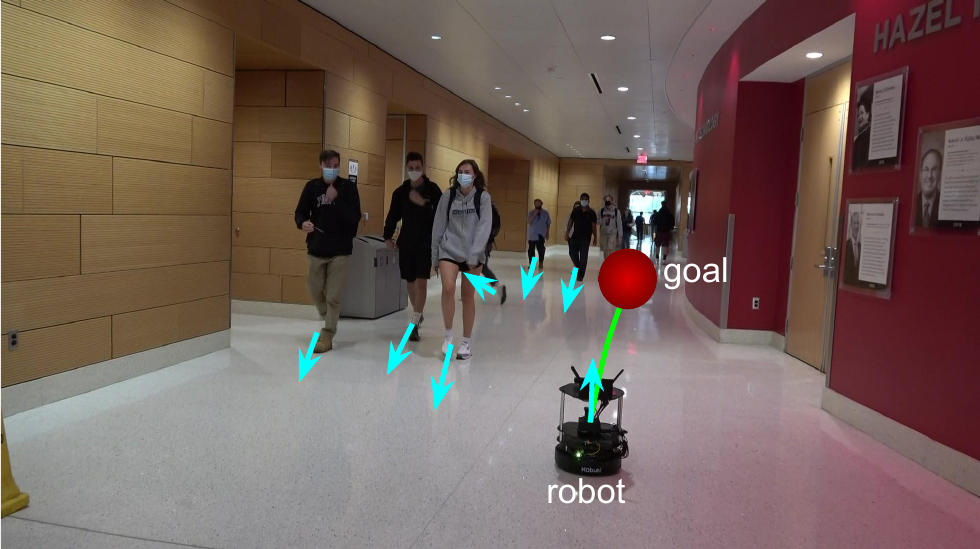}
            \label{fig:hallway_medium_density3}
    }%
    \subfigure[t+3]{
            \centering
            \includegraphics[width=0.235\textwidth]{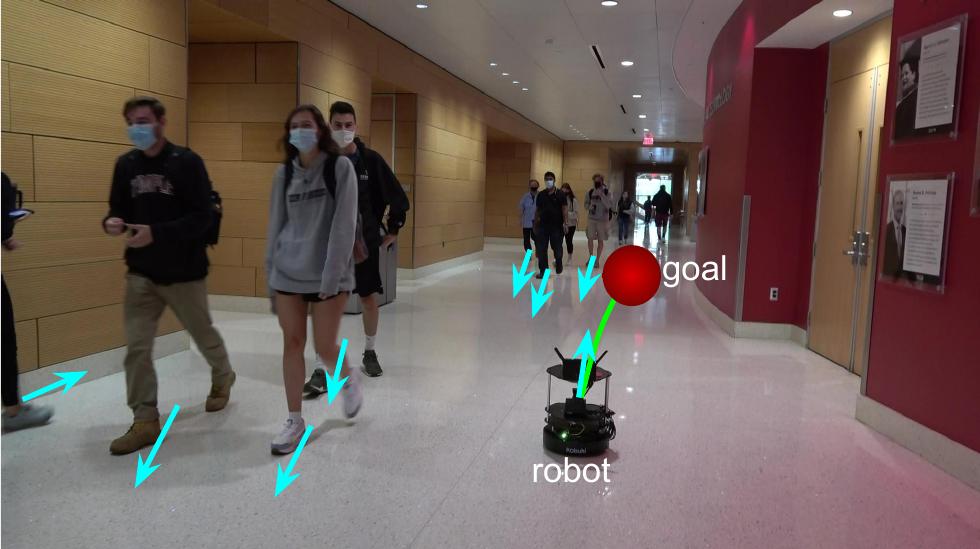}
            \label{fig:hallway_medium_density4}
    }%
    \caption{Robot reactions to moving pedestrians in the indoor hallway with the medium crowd density at different times.
            }
    \label{fig:hallway_medium_density}
\end{figure*}

\begin{figure*}[t]
    \centering
    \subfigure[t]{
            \centering
            \includegraphics[width=0.235\textwidth]{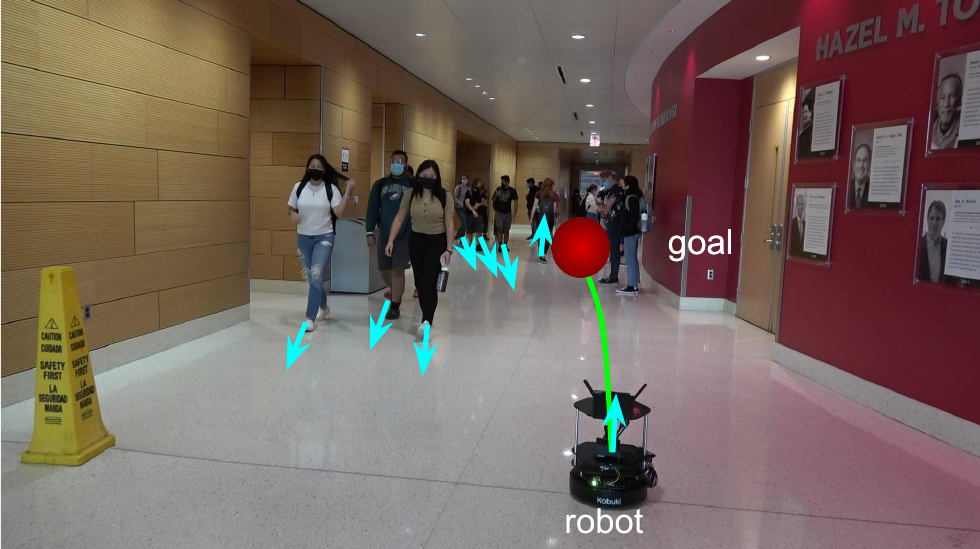}
            \label{fig:hallway_high_density1}
    }%
    \subfigure[t+1]{
            \centering
            \includegraphics[width=0.235\textwidth]{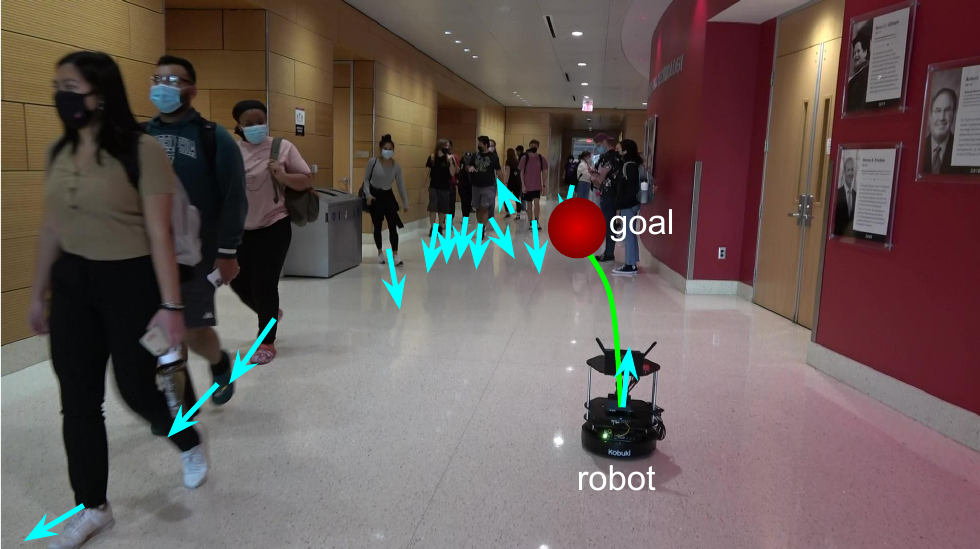}
            \label{fig:hallway_high_density2}
    }%
    \subfigure[t+2]{
            \centering
            \includegraphics[width=0.235\textwidth]{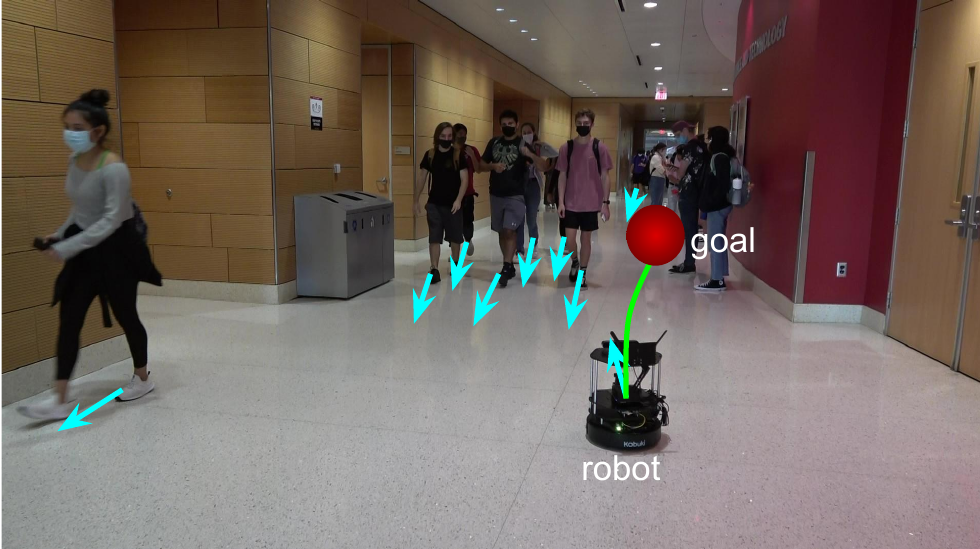}
            \label{fig:hallway_high_density3}
    }%
    \subfigure[t+3]{
            \centering
            \includegraphics[width=0.235\textwidth]{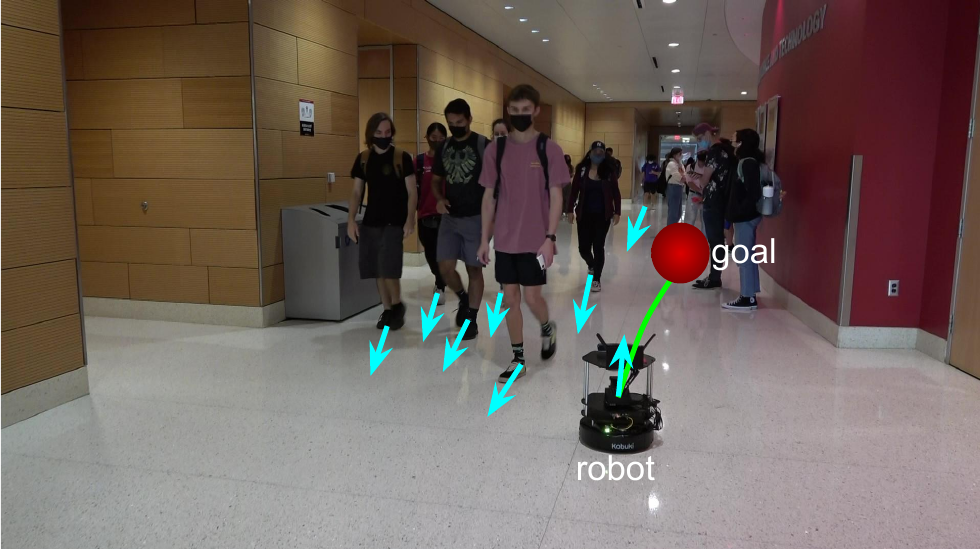}
            \label{fig:hallway_high_density4}
    }%
    \caption{Robot reactions to moving pedestrians in the indoor hallway with the high crowd density at different times.
            }
    \label{fig:hallway_high_density}
\end{figure*}

To test the real-world generalization across different crowd sizes, we let the robot navigate through a sequence of 15 goal points in the indoor hallway shown in Fig.~\ref{fig:indoor_hallway}.
We tested the robot at different times, getting results with different crowd densities.
Figure~\ref{fig:hallway_low_density}-\ref{fig:hallway_high_density} and Multimedia Extension 2 shows the schematic diagrams of robot navigation behavior and experimental videos for low, medium, and high crowd density.

In the low density crowd, Multimedia Extension 2 and Fig.~\ref{fig:hallway_low_density} show how our robot safely drives a total of \unit[131.24]{m} in the nearly empty hallway at an average speed of \unit[0.41]{m/s}. 
It is apparent that in this case, our robot can easily navigate to its goals safely and quickly by tuning its heading direction towards sub-goal points and following the nominal path.
Similarly, our robot is also able to safely and quickly navigate to its goal points through the medium-density crowds, traveling a total of \unit[137.28]{m} at an average speed of \unit[0.43]{m/s} without any collisions, as shown in Multimedia Extension 2 and Fig.~\ref{fig:hallway_medium_density}.
The high density crowd, shown in Multimedia Extension 2 and Fig.~\ref{fig:hallway_high_density}, is more complex than the first two cases as there are stationary pedestrians in addition to the groups of moving pedestrians.
Even in such a complex dynamic environment, our robot is still able to actively avoid collisions with moving students going in or going out of classes, safely avoid static students standing or sitting in the hallway, and quickly reach predefined goals, traveling a total length of \unit[133.40]{m} and an average speed of \unit[0.41]{m/s}.

These results demonstrate that our navigation policy is able to generalize to yet another new environment with varying crowd densities and, more importantly, is able to seamlessly transition from simulation to the real world.
Furthermore, we see that the trend of maintaining a high irrespective of the crowd density is consistent with the simulations in Sec.~ \ref{subsubsec:different_crowd_densities}.

\subsubsection{Different Environments}
\label{subsubsec:different_environments_hardware}
We also investigate the ability of our DRL-VO control policy to generalize across different real-world environments. 
Aside from the indoor hallway environment, we also let the robot navigate through different sequences of 10 goal points in both the indoor lobby and outdoor hallway environments, shown in Fig.~\ref{fig:indoor_lobby} and Fig.~\ref{fig:outdoor_hallway}, respectively.
Note that all these three environments are tested under high crowd density, forming two sets of comparative experiments.
One is the comparison of the structured indoor lobby and the relatively open indoor hallway.
Another is the comparison of the indoor hallway and the outdoor hallway.

Multimedia Extension 3 video and Fig.~\ref{fig:lobby_high_density} show the navigation behavior of our robot in the indoor lobby environment, where the robot successfully navigates through 10 goal points and travels a total of \unit[68.96]{m} at an average speed of \unit[0.37]{m/s}. 
Compared with the relatively open indoor hallway environment, this structured lobby environment is more complex and unpredictable, with static chairs, tables, bicycles and trash bins as well as many dynamic pedestrians, resulting in a narrow and crowded free space.
This structure yielded a more winding nominal path for the robot to follow, which frequently changed as pedestrians cut of different narrow passageways.
Despite these complexities, we found no significant navigation differences between these two environments, and our robot was still able to safely and quickly reach its goals. 
This qualitative comparison demonstrates that our DRL-VO policy is able to generalize to relatively open environments and structured environments in the real world. 

We next analyze robot navigation in the outdoor hallway environment. 
As can be seen from Multimedia Extension 3 and Fig.~\ref{fig:outdoor_high_density}, our robot navigates smoothly in the dense crowds and goes to its predefined goals without any collisions. 
It travels a total of \unit[102.73]{m} at an average speed of \unit[0.43]{m/s} through this open space environment.  
Note that our outdoor hallway is more open than that indoor hallway, with no static obstacles but many more pedestrians walking around. 
Compared with navigating in the indoor hallway, the navigation behavior of our robot again shows no significant observable difference. 
This outdoor experiment indicates that our DRL-VO policy can also work reliably in outdoor environments and it is robust to both indoor and outdoor environments.

One interesting observation from these real-world experiments is the curiosity of the pedestrians, who frequently stopped to observe the robot's behavior, a phenomenon that we did not include in the training data but which the robot was able to easily handle.
We also observed that most people actively avoided the robot when they noticed it, giving it a wide berth.
Recall we did model this behavior by adding an additional social force component to push people away from the robot and that this repulsion force was stronger than that between two pedestrians.

In summary, these real-world experiments and results demonstrate that our DRL-VO policy can smoothly transfer from simulation to reality and be applied to real-world tasks.
We believe there are three main reasons for this. 
First, different from other works using 2D simulators \cite{long2018towards, guldenring2020learning, chen2017socially, everett2018motion, everett2021collision, chen2019crowd, liu2020robot, liu2020decentralized, chen2020relational, dugas2020navrep, xu2021human}, we use a 3D simulator to train our control policy, which shrinks the sim-to-real gap by modeling anthropomorphic 3D pedestrian and precise robot kinematics to give more realistic information than the point pedestrian and robot models.
Second, we preprocessed the sensor data before inputting it into our navigation policy, turning the raw sensor data into the kinematic and lidar grids.
These data structures are less sensitive to the differences between simulation and reality than raw data is (\eg simulated camera images are much different than real images due to the lack of detailed texture in the environment). 
Third, the VO-based heading direction reward function is a dense reward and always motivates the robot to actively move both towards goal points and away from potential collisions.   
All of these factors reduce the gap between simulation and reality, allowing our policy learned entirely in simulation to work in the real world without any retraining or fine-tuning.

\section{Highly Constrained Environments}
\label{sec:static_results}
\begin{figure*}[t]
    \centering
    \subfigure[World 62]{
            \centering
            \includegraphics[width=0.31\linewidth]{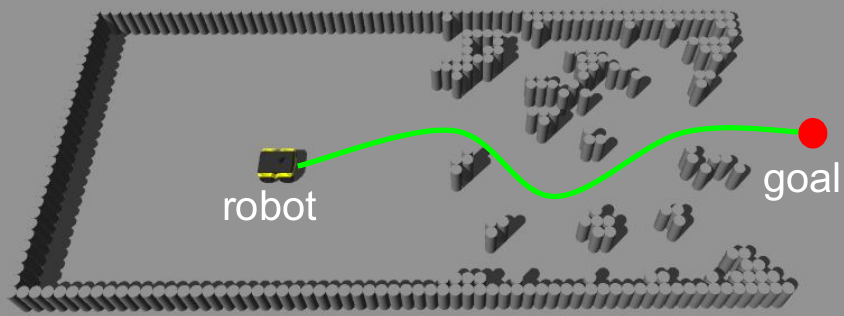}
            \label{fig:barn_62}
    } 
    \subfigure[World 120]{
            \centering
            \includegraphics[width=0.31\linewidth]{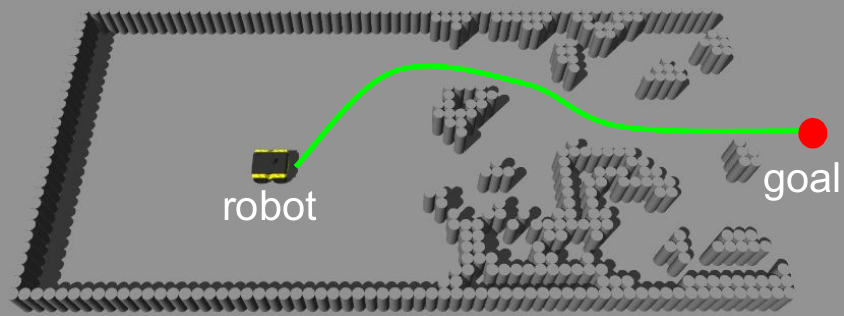}
            \label{fig:barn_120}
    } 
    \subfigure[World 285]{
            \centering
            \includegraphics[width=0.31\linewidth]{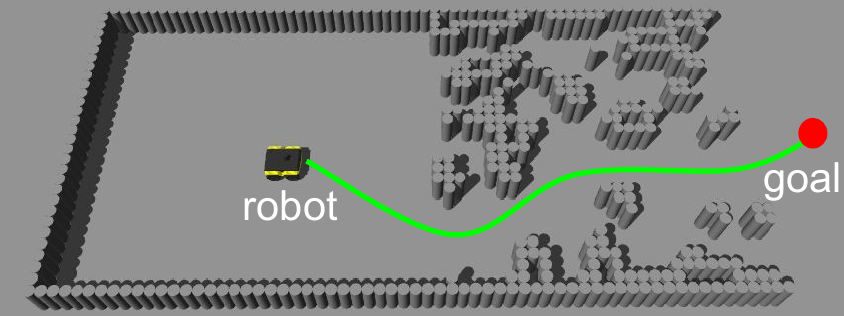}
            \label{fig:barn_285}
    }%
    \caption{Example BARN simulated environments.}
    \label{fig:barn_sim}
\end{figure*}

We further test our policy's ability to generalize to other environments and robot models. To do this, we entered our DRL-VO control policy in the ICRA 2022 BARN Challenge \cite{xiao2022autonomous}, where the goal is to navigate through unknown and highly constrained static environments, such as those shown in Fig.~\ref{fig:barn_sim}.
The competition consisted of two phases, simulation and hardware.

\subsection{Experimental Setup}
\label{subsec:experimental_setup_barn}

\subsubsection{Robot Platform}
The BARN challenge uses a Jackal robot equipped with a Hokuyo UTM-10LX lidar.
Compared to the Turtlebot2 platform used in Sec.~\ref{sec:dynamic_results}, the maximum velocity of the Jackal is much greater (\unit[2]{m/s} vs.\ \unit[0.5]{m/s}) and the sensing range and angular resolution of the UTM-10LX are both smaller (\unit[10]{m} and $0.375^\circ$ per beam vs.\ \unit[30]{m} and $0.25^\circ$ per beam, respectively).

\subsubsection{Simulation}
\label{subsec:simulation_configuration_barn}
The simulation portion of the challenge used 300 publicly available training Gazebo environments and 50 test environments known only to the competition managers. Each map is filled with cylindrical obstacles generated by the BARN environment generator \cite{perille2020benchmarking}, and the object density varies greatly across environments. The competition evaluated all policies on a desktop computer with an Intel Xeon Gold 6342 CPU @ 2.80 GHz (without GPU support).

\subsubsection{Hardware}
\label{subsec:hardware_configuration_barn}
\begin{figure*}[t]
    \centering
    \hfill
    \subfigure[Wide ``S'' shaped environment]{
            \centering
            \includegraphics[width=0.31\linewidth]{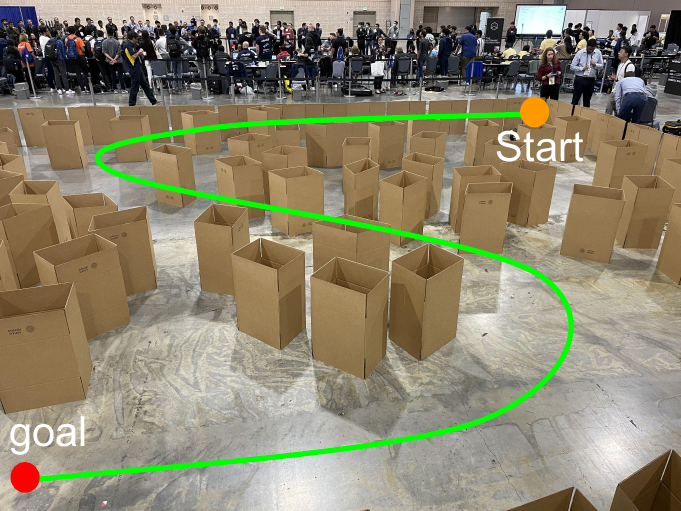}
            \label{fig:barn_real1}
    } \hfill
    \subfigure[Pointed ``S'' shaped environment]{
            \centering
            \includegraphics[width=0.31\linewidth]{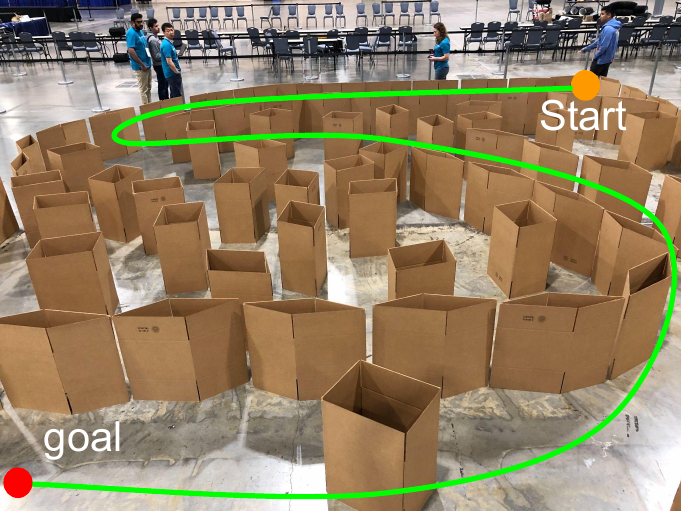}
            \label{fig:barn_real2}
    } \hfill
    \subfigure[Narrow ``S'' shaped  environment]{
            \centering
            \includegraphics[width=0.31\linewidth]{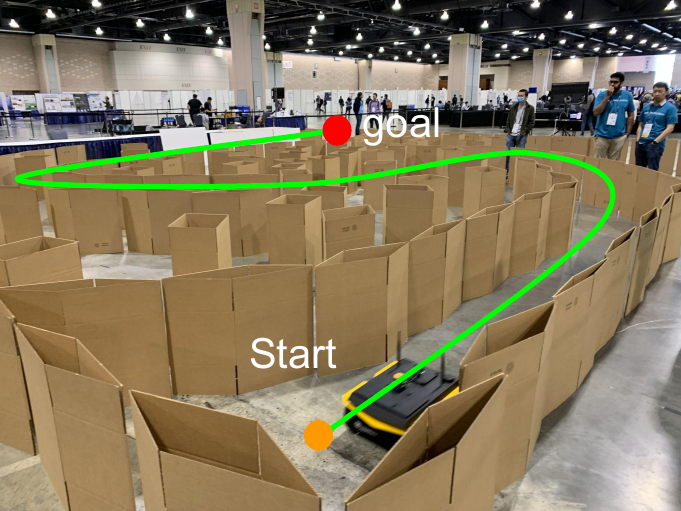}
            \label{fig:barn_real3}
    } \hfill
    \caption{Real-world BARN test environments.}
    \label{fig:barn_real}
\end{figure*}

There were three different test environments, each about \unit[$10\times10$]{m} in size: wide ``S'', pointed ``S'', and narrow ``S''.
Unlike the semi-closed BARN simulation environments, these three physical test environments are complex mazes with only one reasonable path, many sharp turns, gap traps, and narrow passages, as Fig.~\ref{fig:barn_real} shows.
The physical Jackal robot was equipped with an onboard computer with an Intel i3-9100TE CPU and \unit[16]{GB} RAM memory (with no GPU) running Ubuntu 18.04 and ROS Melodic Morenia.

\subsection{Deployment Adjustment}
\label{subsec:deployment_adjustement}
We utilized the same DRL-VO policy from Sec.~\ref{sec:training_procedure} (trained using the Turtlebot2) in the BARN Challenge.
However, to successfully deploy policy on the Jackal robot and meet the requirements of the BARN challenge, we needed to adjust several settings of our overall system.

First, the BARN challenge maps are all unknown to the robot. Therefore, we removed the localization module (\ie AMCL) and adapted our navigation system to operate using only odometry (\ie without a map).

Second, we need to adjust the input observation data $\mathbf{o}^t = [\mathbf{l}^t, \mathbf{p}^t, \mathbf{g}^t]$ of the DRL-VO network.
For the lidar data $\mathbf{l}^t$, since the UTM-10LX has only 720 scan points instead of 1080 scan points, we use the full $270^\circ$ FOV (instead of the $180^\circ$ FOV for the Turtlebot2) to construct the $80 \times 80$ lidar historical map $\mathbf{l}^t$.
We set the pedestrian data $\mathbf{p}^t = 0$ since the maps only contain stationary obstacles.
Lastly, we use a sub-goal point with a look-ahead distance of \unit[1]{m} instead of \unit[2]{m} to generate the goal $\mathbf{g}^t$ as this allows the robots to better avoid obstacles in the highly constrained environments.

Finally, we found that directly mapping the normalized velocity from the control policy to the speed range of the Jackal led to very aggressive behavior. We hypothesize that this is an artifact of the Turtlebot2 being slower than many pedestrians so it nearly always moves at (close to) maximum speed. However, this caused the Jackal to frequently bump into obstacles. To solve this problem, we designed a simple maximum velocity switching mechanism:
\begin{equation}
    v_{\rm max} =
    \begin{cases}
    \unit[2]{m/s} & \text{if $d^t_{\rm obs} \geq d_f$} \\
    \unit[0.5]{m/s} &\text{otherwise},
    \end{cases}
    \label{eq:v_max}
\end{equation}
where $v_{\rm max}$ is the maximum velocity limit of the robot, $d^t_{\rm obs}$ is the distance to the obstacle closest to the robot at time $t$, and $d_f (= \unit[2.2]{m})$ is a threshold value.
Intuitively, this equation sets the maximum velocity to \unit[2]{m/s} only when there are no obstacles in front of the robot and uses a slower velocity near obstacles (matching the training velocity).
We open-sourced the DRL-VO policy code with these adjustments at \url{https://github.com/TempleRAIL/nav-competition-icra2022-drl-vo}.

\subsection{Results}
We competed in both the simulation and hardware phases of the BARN competition.

\subsubsection{Simulation}
\label{subsec:simulation_results_barn}
The BARN challenge simulation competition uses 50 unseen Gazebo environments to test navigation policies, running 10 trials for each environment.
All trials use the same start and goal points, with the goal \unit[10]{m} in front of the start, which allows the robot to plan a reasonable nominal path in the absence of an environmental map. Policies were scored using a metric $s$ that considers the success rate, travel time, and environment difficulty and yields a value in the range $[0, 0.25]$ (where 0.25 is best) \cite{xiao2022autonomous}.

A total of 11 policies (\eg model-based, learning-based, and hybrid) participated in the BARN challenge simulation competition.
Our DRL-VO policy has the average highest navigation score among all policies at 0.2415, and achieved a score of 0.25 on all maps that it successfully completed (meaning the average speed was at least \unit[0.5]{m/s}).
The other competitors' scores ranged from 0.1627 to 0.2334.
See Multimedia Extension 4 to see our policy in these highly constrained environments.
We did not receive full data about the test environments, but based on our experience in the training environments we saw that failures happened when the robot would just barely bump into an obstacle near very small gaps and sharp turns.
We believe that this is because the Jackal has a larger physical size than the Turtlebot2, so the policy likely would have been safe with a Turtlebot2.

\begin{figure*}[t]
    \centering
    \subfigure[t]{
            \centering
            \includegraphics[width=0.235\textwidth]{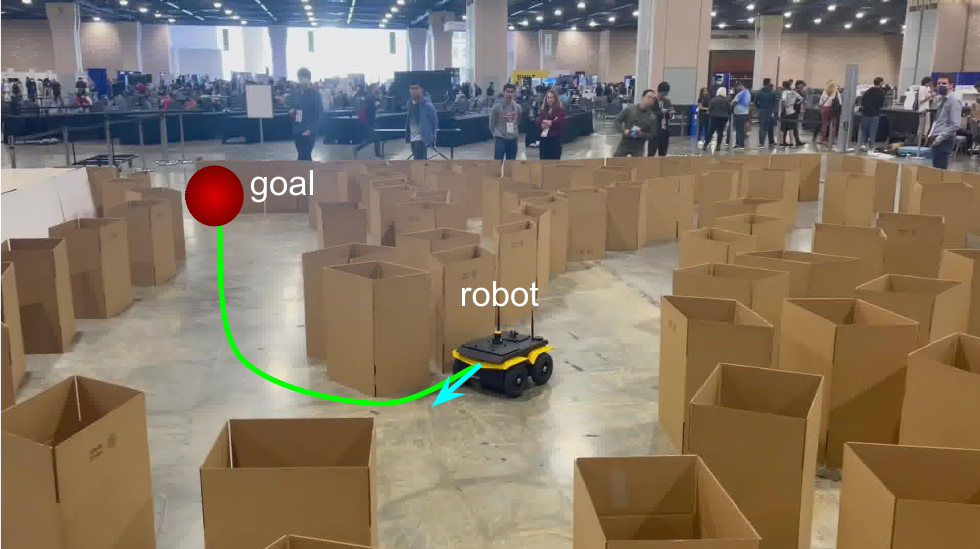}
            \label{fig:barn_real_success1}
    }%
    \subfigure[t+1]{
            \centering
            \includegraphics[width=0.235\textwidth]{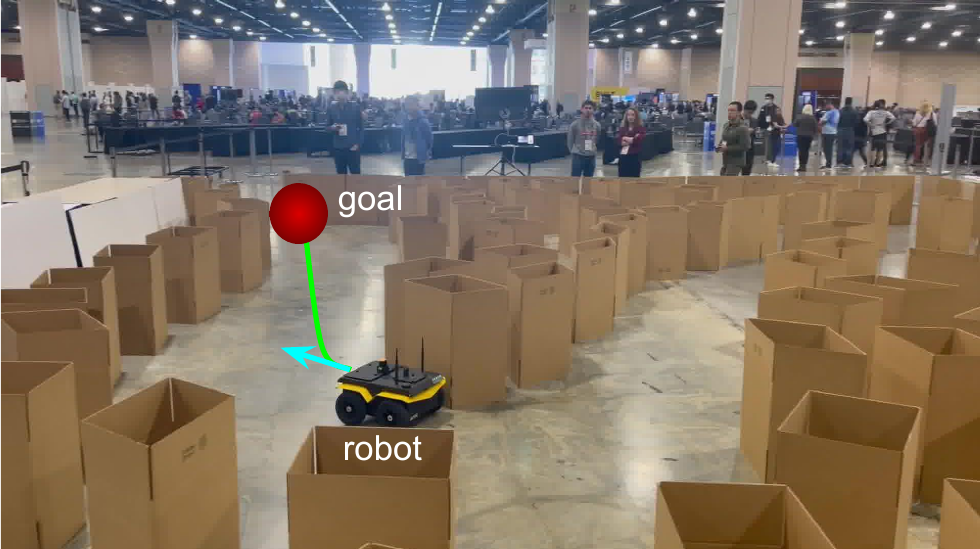}
            \label{fig:barn_real_success2}
    }%
    \subfigure[t+2]{
            \centering
            \includegraphics[width=0.235\textwidth]{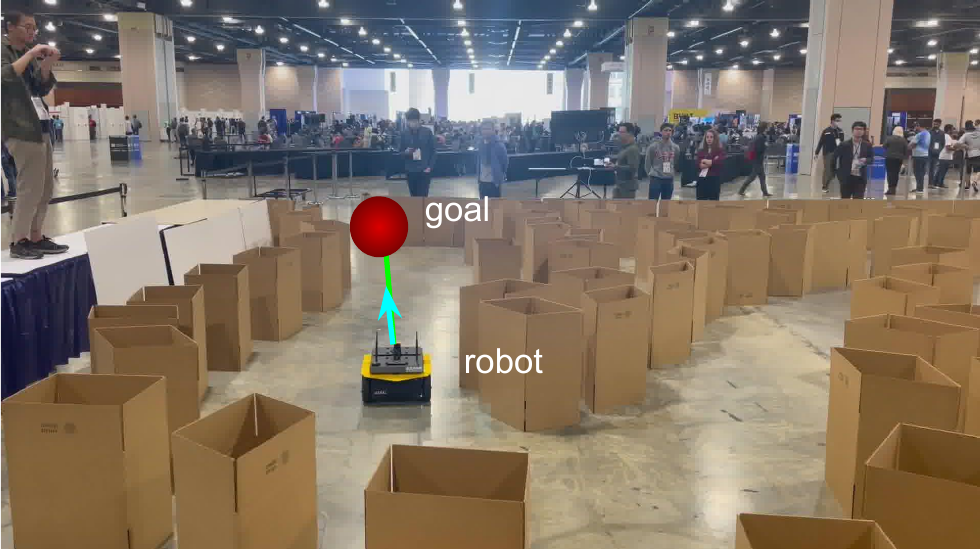}
            \label{fig:barn_real_success3}
    }%
    \subfigure[t+3]{
            \centering
            \includegraphics[width=0.235\textwidth]{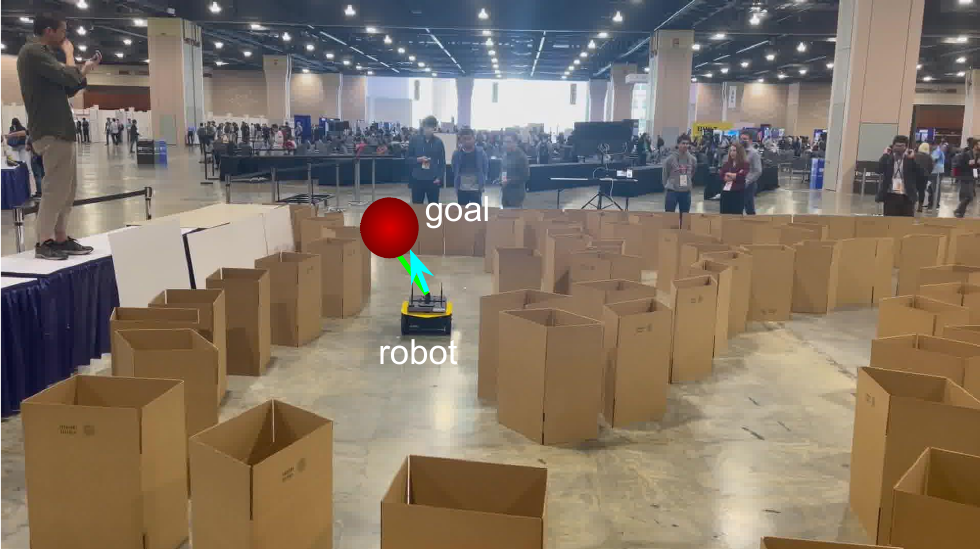}
            \label{fig:barn_real_success4}
    }%
    \caption{Successful navigation behavior of the Jackal robot in the wide ``S'' shaped environment at different times.
            }
    \label{fig:barn_success}
\end{figure*}

\subsubsection{Hardware}
\label{subsec:hardware_results_barn}
Due to the excellent navigation performance achieved by our DRL-VO policy in the BARN simulation competition, we were invited to participate in the physical competition (along with two other teams, ARML and DRAGON).
The structure of the physical competition was to conduct three runs in each of the three test environments shown in Fig.~\ref{fig:barn_real}. 
The team that completed the most runs (out of 9 total) won, with the average time used a tiebreaker. The test environments were generated by the competition managers that day and were not known to competitors beforehand. 
As we can see, these environments are more challenging than the simulation test environments, with many sharp turns and gap traps where the robot's field of view is highly occluded.
Additionally, some of the passageways were so narrow that the Jackal robot could not turn around without hitting a wall.

\begin{figure}
    \centering
    \subfigure[Sharp turns]{
            \centering
            \includegraphics[width=0.3\linewidth]{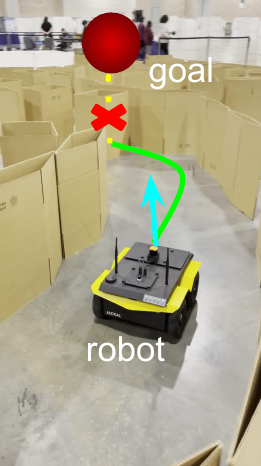}
            \label{fig:barn_real_failure1}
    }%
    \subfigure[Gap traps]{
            \centering
            \includegraphics[width=0.3\linewidth]{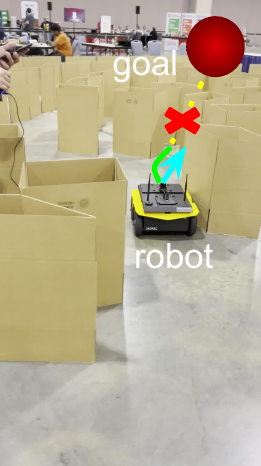}
            \label{fig:barn_real_failure2}
    }%
    \subfigure[Narrow passages]{
            \centering
            \includegraphics[width=0.3\linewidth]{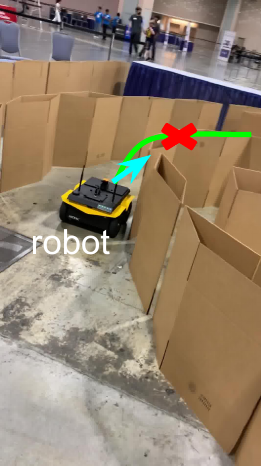}
            \label{fig:barn_real_failure3}
    }%
    \caption{Typical failure cases in the pointed/narrow ``S'' shaped environment. 
    The yellow dotted lines are the unreasonable nominal paths given by the odometry-based global planner due to the limited FOV of the robot. 
    The red crosses indicate potential collision areas.}
    \label{fig:barn_failure}
\end{figure}

We placed third in the physical competition, completing 2 out of 9 runs successfully (both in the first environment, Fig.~\ref{fig:barn_real1}).
Multimedia Extension 5 and Fig.~\ref{fig:barn_success} show one of these successful runs through the first test environment (Fig.~\ref{fig:barn_real1}), which had relatively wide passages.
However, we were not able to complete a full run in either of the other two environments due to two primary reasons: 1) there were no reasonable nominal paths to guide the robot through sharp turns or gap traps due to unpredictable obstacles limiting the robot's view, and 2) our DRL-VO policy is a bit aggressive and lacks precise control directives in the narrow passages, as it is trained in an environment without narrow passages and on a robot with a smaller physical size.
See Extension 5 and Fig.~\ref{fig:barn_failure} for examples of these failures.
In the future, we plan to alleviate these issues by using a map-based navigation system to give a better nominal global path to the robot and by training our DRL-VO policy in multiple different environments.
Additionally, relative to the model-based approaches used by our competition, we could quickly retrain or fine-tune our learning-based model to work in each new environment.

\subsection{Discussion}
Participating in the simulated and hardware BARN competition showed that our DRL-VO control policy, which was trained in a single crowded dynamic Lobby environment with 34 pedestrians using a Turtlebot2 robot, was able to generalize to a new navigation task (\ie highly constrained static environments) and work with different robot platforms and sensor configurations.
We believe that this generalizability can be explained by the fact that we use preprocessed data representations (Sec.~\ref{subsec:obervation_space}) rather than the raw sensor data. This creates a level of abstraction between the hardware/environment and the policy that allows it to be highly flexible.
We will continue to explore the transferability of our policy in future work, particularly looking at how to account for changes in the physical properties of a robot.
Two examples that we encountered in this work were the differences in the physical size of a robot (which we believe led to many of the collisions in the narrow passageways) and in the speed of the robot (since a Jackal robot moving in a crowd of people should vary its speed based on the people rather than going full speed).

\section{Lessons Learned} 
\label{sec:guidelines}
Neural networks are, in essence, complicated tools to approximate functions.
As such, they are particularly well suited to solve problems that are too complex for traditional rule-based methods, such as control policies to enable robots to navigate through dense crowds of pedestrians.
There are three primary aspects to a neural network: the input data, the network structure, and the training methodology.
In this section, we will discuss some of the broad lessons we have learned from our work in each of the above aspects.

\subsection{Data Representations}
\label{subsec:data_preprocessing_and_representation}
We argue that creating intermediate data representations (\ie hand-crafted features or learned features) from the raw sensor data allows us to learn navigation policies that better generalize to new scenarios than policies that directly use raw sensor data alone.
As we conclude in Sec.~\ref{subsec:summary_analysis}, most robot navigation methods (\ie model-based \cite{fox1997dynamic, singamaneni2021human, fiorini1998motion, wilkie2009generalized, van2011reciprocal, arul2021v, brito2019model, shen2020collision} and learning-based \cite{chen2017socially, everett2018motion, everett2021collision, chen2019crowd, chen2020robot, liu2020decentralized, liu2020robot, sathyamoorthy2020densecavoid, chen2020relational, dugas2020navrep, pateldwa}) utilize the preprocessed intermediate data as the input to their control policies, with most of the learning-based control policies also utilizing learned intermediate features.
However, learning useful intermediate features from raw sensor data requires the use of photorealistic simulators and/or real data, both of which present significant challenges during the learning phase.
Additionally, even photorealistic simulators differ (subtly) from reality, which can create a sim-to-real gap as the learned intermediate features depend on the data used to train the network.

Instead, we carefully handcrafted features in an attempt to concisely summarize the salient information needed for navigation while abstracting away the complexities of the real world.
In our case, we choose two inputs to encode relevant data about the structure of the environment and the relative kinematics of pedestrians (or other moving objects).
Specifically, we compress the lidar data by using minimum and average pooling to encode information about the basic structure of the space.
This step allows our policy to avoid overfitting to the specific training environment, based on the findings of \cite{pfeiffer2018reinforced}.
Similarly, we encode the RGB-D camera data by detecting objects from the RGB value, measuring positions using the depth data, and fusing these in the MHT to get information about pedestrian kinematics.
Overall, these preprocessing steps form a compact and useful representation that is far less sensitive to pixel-level differences between individual images than policies that use raw sensor data as input.
This choice also affects the network design and training methodology, as we will discuss below.

\subsection{Computational Constraints}
\label{subsec:computational_constraints}
The goal of much machine learning research is simply to improve accuracy and/or reward.
However, when using machine learning in the context of robotics it is critical to remember that the resulting algorithm will be deployed on a mobile robot with limited computational resources.
Therefore, in practice, one must often make a trade-off between accuracy/reward and computational efficiency.
Based on our experience, we believe that having a short reaction time is more important than using a policy that is incrementally better (in the sense of lower training loss or higher reward in a simulator that runs slower than real-time) but that runs much slower and cannot be processed in real-time.
In other words, we believe one should attempt to maximize reward subject to a constraint on processing time rather than the other way around.
Future work will study this hypothesis in greater detail.

\subsection{Training Methodology}
In a deep reinforcement learning setting, there are two key aspects to training: the environment and the reward design.

\subsubsection{Training Environment}
In general, reinforcement learning must take place in simulation due to the large number of iterations needed for the policy to converge and the need to explore potentially unsafe regions of the control space before convergence.
One direction of work in the navigation research community is to develop photorealistic simulators \cite{makoviychuk2021isaac}.
We agree that these are necessary to learn control policies that use raw images as their input.
However, our input data structures provide a level of abstraction between the visual scene and the control policy by encoding the detail (or lack thereof) in the world in a consistent set of representations.
This allows us to avoid using a computationally complex simulator and instead have an environment with flat visual textures and identical people-shaped rigid bodies that float along the ground.
As a result, each iteration takes less time while still effectively learning a policy that generalizes well to new environments, including going from a simulated world to the real world without the need for any modifications.

Another important aspect of the training environment is the realism of human behavior.
There are three types of pedestrian crowd models used by navigation-related researchers: social force-based model \cite{ferrer2013robot, guldenring2020learning, xie2021towards, singamaneni2021human}, velocity obstacle-based model \cite{ perez2020robot, chen2017socially, everett2018motion, everett2021collision, liu2020robot, liu2020decentralized, chen2019crowd, chen2020relational, dugas2020navrep, sathyamoorthy2020densecavoid}, and pedestrian dataset-based model \cite{sebastian2017socially, chen2020robot, xu2021human}.
These works, especially those driven by pedestrian datasets, only modeled pedestrian dynamics and completely ignore any human-robot interactions.

However, during our real-world experiments, we observed three distinct types of human reactions to the robot, each of which affect the robot in a different way.
The first, and most common, was for pedestrians to slightly adjust their path to avoid the robot but to continue on their way.
Second, some pedestrians were curious about our robot and would often stop and watch or follow the robot to observe its behavior.
Third, and least commonly, some pedestrians were afraid of the robot and would try to get away from the robot quickly. This list is by no means exhaustive, and there are likely many more human-robot interactions that affect navigation.
Instead, this list is meant to demonstrate that approaches that completely ignore human reactions to robots will be less successful and have a larger gap between simulation and reality.
Our approach, also used by several other groups \cite{ferrer2013robot, guldenring2020learning}, is a simplistic one: add an additional social force to repel people and robots away from one another.
We believe that one fruitful direction for future work will be to explore the range of human behaviors, determine what effect they have on the robot, and develop methods to (approximately) reconstruct this behavior in a simulated setting.

\subsubsection{Reward Design}
The question of how to design an effective reward function for a specific application is still an open problem.
For our robot navigation application, we want the prioritize collision avoidance while still reaching the goal in the shortest time possible.
As we previously discussed, most DRL-based navigation policies use sparse reward functions, including a zeroth-order (\ie distance-based) collision avoidance term and a goal-attainment term, which guide the robot to avoid collisions and reach the goal, respectively.
However, these zeroth-order collision avoidance terms force the robot to reactively avoid collisions instead of proactively taking evasive action.
This was the main motivating factor in using velocity obstacles to create a first-order collision avoidance term, as the velocity information guides the robot to pick actions that will be safe both now \emph{and in the future}.
Our experiments on different crowd densities and unseen environments show that our novel first-order active reward function greatly improves the safety and speed of robot navigation compared to the zero-order passive reward function (DRL-VO vs.\ DRL in Tables~\ref{tab:densities}--\ref{tab:environments35}).
Based on these results, another avenue for future exploration will be designing other higher-order reward functions to improve robots' ability to proactively avoid collisions in highly dynamic scenes.

A secondary motivation for using velocity obstacles was to allow the robot to indirectly learn from successful model-based approaches.
For example, controllers based on velocity obstacles can guarantee collision avoidance under the assumption of perfect knowledge of the environment.
However, in practice, that assumption is never valid.
Therefore, instead of directly applying velocity obstacles to generate control actions, we use it as a reward term to guide behavior that is (at least close to) safe.
Another avenue of future work will be to examine the effectiveness of using other model-based control approaches to guide the design of similar reward terms that combine safety and progress towards the goal into a single term.

\section{Conclusion}
\label{sec:conclusion}
In this paper, we proposed the DRL-VO control policy to enable autonomous robot navigation in crowded dynamic environments with both static obstacles and dynamic pedestrians, as well as in highly constrained static environments.
From the detailed background material analysis, our DRL-VO control policy has two key novelties.
First, we propose a new combination of preprocessed data representations, which can work well in crowded dynamic environments with up to 55 pedestrians and bridge the appearance gap between an imperfect simulation and reality.
Specifically, the robot fuses a short history of lidar data, current pedestrian kinematics (\ie position and velocity stored in the two grids), and a sub-goal point.
All of this data is represented within the robot's local coordinate frame, making our navigation policy robust to errors in localization that are common in crowded, dynamic environments.
Second, we design a novel velocity obstacle-based reward function used to train the policy, which can be easily applied to multi-robot navigation policies.
This reward uses first-order information about pedestrians to guide the robot to actively steer towards the goal while avoiding potential future collisions.

We demonstrate that our DRL-VO policy generalizes better and maintains a better balance between collision avoidance and speed to different crowd sizes and different unseen environments than other state-of-the-art model-based, supervised learning-based, and DRL-based policies, achieving an almost constant average speed but with a higher success rate over a series of 3D simulation experiments.
We also demonstrate through extensive hardware experiments that our DRL-VO policy overcomes the sim-to-real gap and works reliably in real-world indoor and outdoor environments with different crowd sizes, and that it is able to do so without any fine-tuning or modification.
Furthermore, by participating in the ICRA 2022 BARN simulation and real-world challenge competitions and placing 1st and 3rd respectively, we demonstrated that our DRL-VO policy can also generalize well to highly constrained environments, different input sensors, and different types of robots.
Finally, we summarize several lessons learned from our work to help other researchers improve the generalizability of learning-based algorithms, choose network architectures for use on mobile robot platforms, bridge the gap between simulation and reality, and design appropriate reward functions.

\section*{Acknowledgment}
\label{sec:acknowledgment}
This research includes calculations carried out on HPC resources supported in part by the National Science Foundation through major research instrumentation grant number 1625061 and by the US Army Research Laboratory under contract number W911NF-16-2-0189.

\bibliographystyle{IEEEtran}
\bibliography{bib/IEEEabrv,bib/self,bib/refs}
\vspace{-15mm}

\begin{IEEEbiography}[{\includegraphics[width=1in,height=1.25in,clip,keepaspectratio]{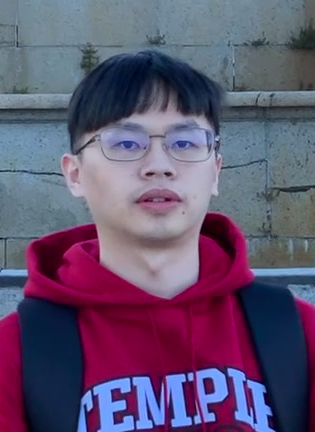}}]{Zhanteng Xie}
    received the B.Eng. degree in electronic information engineering from Zhengzhou University, Zhengzhou, China, in 2015 and the M.Eng. degree in information and communication engineering from the Harbin Institute of Technology, Harbin, China, in 2018. He is currently working toward the Ph.D. degree with the Temple Robotics and Artificial Intelligence Lab (TRAIL), Department of Mechanical Engineering, Temple University, Philadelphia, PA, USA.
    
    His research interests lie at the intersection of robotics and machine learning, with a focus on environment prediction and autonomous robot navigation in crowded dynamic scenes.
\end{IEEEbiography}
\vspace{-15mm}

\begin{IEEEbiography}[{\includegraphics[width=1in,height=1.25in,clip,keepaspectratio]{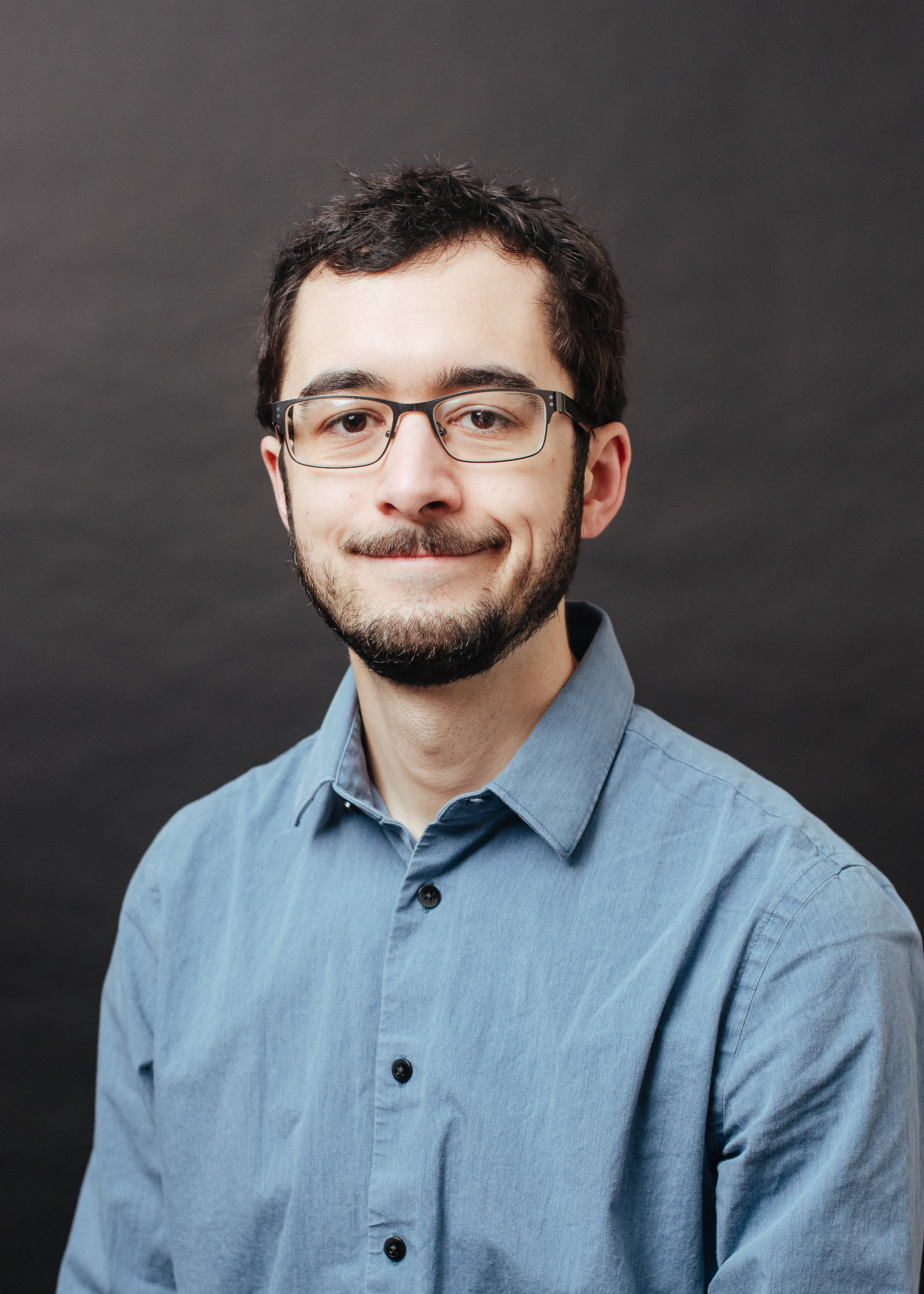}}]{Philip Dames}
    received both his B.S. (summa cum laude) and M.S. degrees in mechanical engineering from Northwestern University, Evanston, IL, USA, in 2010 and his Ph.D. degree in mechanical engineering and applied mechanics from the University of Pennsylvania, Philadelphia, PA, USA, in 2015. 
    
    From 2015 to 2016, he was a Postdoctoral Researcher in Electrical and Systems Engineering at the University of Pennsylvania. Since 2016, he has been an Assistant Professor of Mechanical Engineering with Temple University, Philadelphia, PA, USA, where he directs the Temple Robotics and Artificial Intelligence Lab (TRAIL). His research aims to improve robots’ ability to operate in complex, real-world environments to address societal needs.

    He is the recipient of an NSF CAREER award.
\end{IEEEbiography}

\enlargethispage{-9.5cm}

\end{document}